\documentclass[10pt,twocolumn,letterpaper]{article}

\usepackage{cvpr}              % To produce the CAMERA-READY version
\usepackage[dvipsnames]{xcolor}

\definecolor{ben}{rgb}{0.9,0.,0.5}

\newcommand{\OURS}{ParDy-Human}

\definecolor{cvprblue}{rgb}{0.21,0.49,0.74}
\usepackage[pagebackref,breaklinks,colorlinks,citecolor=cvprblue]{hyperref}
\usepackage{multirow}
\usepackage{hyperref}

\def\paperID{12095} % *** Enter the Paper ID here
\def\confName{CVPR}
\def\confYear{2024}

\title{Deformable 3D Gaussian Splatting for Animatable Human Avatars}
\author{
HyunJun Jung$^{1}$, 
Nikolas Brasch$^{1}$,
Jifei Song$^{2}$, 
Eduardo Pérez-Pellitero$^{2}$, 
Yiren Zhou$^{2}$,
Zhihao Li$^{2}$,\\
Nassir Navab$^{1}$,
Benjamin Busam$^{1,3}$\\[0.5em]
$^1$ Technical University of Munich\qquad
$^2$ Huawei Noah’s Ark Lab \qquad
$^3$ 3dwe.ai\\[0.5em]
{\tt\small \{hyunjun.jung,b.busam\}@tum.de}
\vspace{-0.5cm}}

\begin{document}

\twocolumn[{%
\renewcommand\twocolumn[1][]{#1}%
\maketitle
\begin{center}
    \captionsetup{type=figure}
    \includegraphics[width=\linewidth]{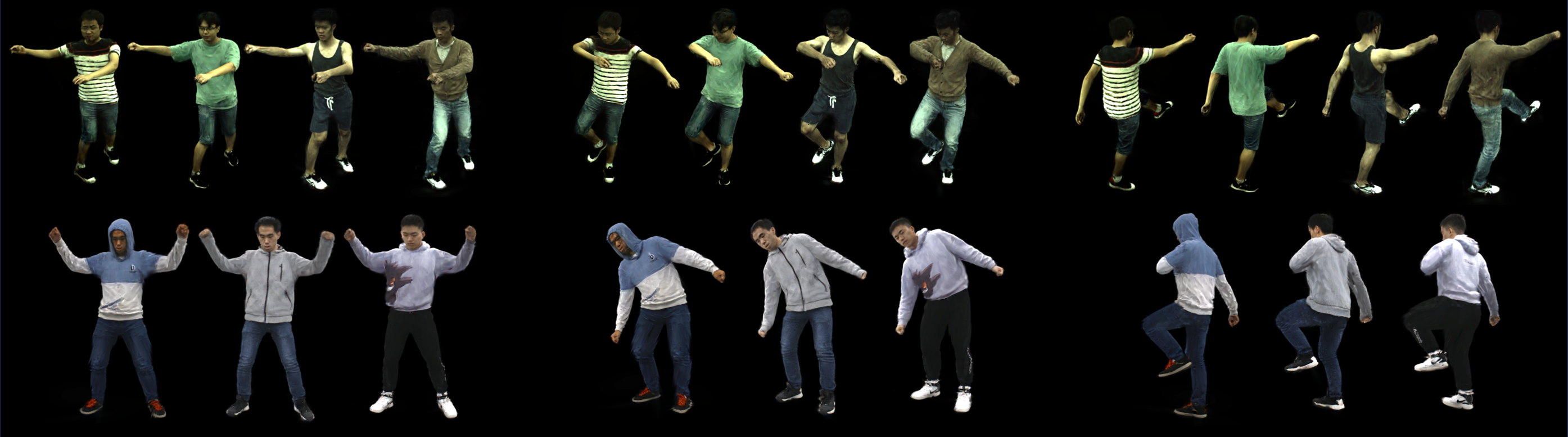}
    \vspace*{-6mm}
    \captionof{figure}{\textbf{\OURS}~constitutes an explicit dynamic human avatar that can be re-posed via SMPL~\cite{loper2015smpl} parameters. It utilizes the design of a deformable version of 3D Gaussian Splatting~\cite{kerbl20233d}. \OURS, unlike existing implicit methods, can be trained with significantly fewer camera views and less human poses. While being free of ground truth mask for training, it generalizes well to novel human poses as shown in the above reposed results on the individuals from ZJU-MoCap~\cite{peng2021neural} and THUman4.0~\cite{zheng2022structured} datasets.}
    \label{fig:teaser}
\end{center}%
}]

\maketitle
\begin{abstract}
Recent advances in neural radiance fields enable novel view synthesis of photo-realistic images in dynamic settings, which can be applied to scenarios with human animation. Commonly used implicit backbones to establish accurate models, however, require many input views and additional annotations such as human masks, UV maps and depth maps. In this work, we propose \OURS~(Parameterized Dynamic Human Avatar), a fully explicit approach to construct a digital avatar from as little as a single monocular sequence. \OURS~introduces parameter-driven dynamics into 3D Gaussian Splatting where 3D Gaussians are deformed by a human pose model to animate the avatar.
Our method is composed of two parts:
A first module that deforms canonical 3D Gaussians according to SMPL vertices and
a consecutive module that further takes their designed joint encodings and predicts per Gaussian deformations to deal with dynamics beyond SMPL vertex deformations.
Images are then synthesized by a rasterizer.
\OURS~constitutes an explicit model for realistic dynamic human avatars which requires significantly fewer training views and images.
Our avatars learning is free of additional annotations such as masks and can be trained with variable backgrounds while inferring full-resolution images efficiently even on consumer hardware.
We provide experimental evidence to show that \OURS~outperforms state-of-the-art methods on ZJU-MoCap and THUman4.0 datasets both quantitatively and visually. Our code is available at \footnotesize{\url{https://github.com/Junggy/pardy-human}}.
\end{abstract}    

\begin{figure*}[!t]
 \centering
    \includegraphics[width=1.0\linewidth]{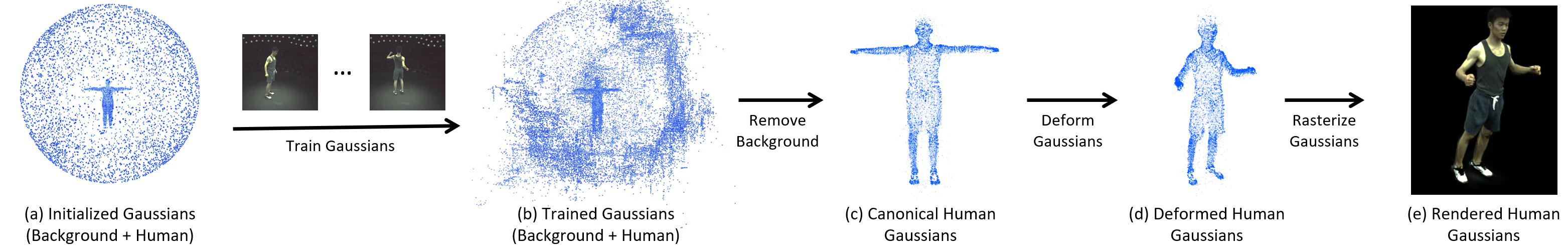}
    \vspace*{-5mm}
    \caption{\textbf{Overview of Avatar Generation Framework.} (a) \OURS~starts by initializing Gaussians on a sphere for the background and a canonical SMPL~\cite{loper2015smpl} mesh for the human. (b) the Gaussians are updated during the training. (c) For inference, the background Gaussians are removed leaving only the avatar. (d) the Canonical human Gaussians are deformed according to the SMPL vertex deformations and learned residual refinement. (e) the deformed Gaussians are rasterized to synthesize an image output under a given pose.} 
    \label{fig:gaussian_overview}
\end{figure*}

\section{Introduction}
\label{sec:intro}
Creating animatable human avatars from images and videos is a popular tasks in computer vision and graphics due to its applications in animation, virtual reality and gaming. Early works~\cite{habermann2020deepcap, alldieck2019learning, saito2021scanimate} are developed around a CNN or MLP based generator that learns 3d human mesh prediction conditioned on a parameterized representation of humans.
While providing a mesh, early approaches produce avatars with only low frequency details for both shape and texture.
Recent works utilize implicit representations combined with volume rendering~\cite{chen2023uv, li2023posevocab, peng2021animatable} which enables photo-realistic rendering with a significantly higher amount of details at the cost of denser camera views and human poses during training.

Most recently, Kerbl \etal~\cite{kerbl20233d} propose an explicit method utilizing point based rendering with 3D Gaussians to enable real-time synthesis of static scenes at high quality.
Other scholars~\cite{luiten2023dynamic} show its potential in dynamic scenarios involving humans.

In this paper, we present \OURS{}~(Parameterized Dynamic Human Avatar), a novel approach to leverage 3D Gaussian Splatting for avatar generation.
We achieve this by updating the canonical T-pose Gaussians during training via posing and deposing with the help of SMPL parameters~\cite{loper2015smpl} and a specialized deformation module.
We design human motion as two consecutive deformations, a naked body deformation followed by motion related cloth movement.

Canonicalized T-posed human Gaussians are deformed via SMPL parameters as a naked body deformation and a MLP associated with the joint's geometric distances determines the residual deformation corrections arising from garment motion.
We initialize the scene with two set of point clouds, one for canonical SMPL vertices representing the human and one randomly covering a spherical surface representing the background.
This split allows the separation between the human and the background, enabling mask-free training while being capable of removing the background at inference time.

We show that \OURS~can be trained without a segmentation mask using significantly fewer cameras and images compared to existing methods.
Experiments proof the realism of rendered avatars in novel poses even with as little as a single monocular video.
We enable image synthesis of full-resolution even on consumer-grade hardware.

%In summary, our contributions can be listed as follows:
In summary, we contribute:

\begin{enumerate}
\item A method for \textbf{deformable 3D Gaussian splatting of human avatars}, a parametrized fully explicit representation for \textbf{dynamic animation} of humans that requires significantly \textbf{fewer views and images} for training.
\item \OURS~provides a \textbf{mask-free training} approach that automatically separates background and human.
\item An efficient pipeline that allows \textbf{significantly faster inference} for \textbf{full-resolution renderings}.% even on a laptop.
\end{enumerate}

\section{Related Work}
\label{sec:related_work}
\subsection{View Synthesis of 3D Scenes}
3D scenes are represented in a plethora of different ways. Classical explicit representations include point clouds, voxels, and meshes, while classical \textbf{implicit representations} utilize signed distance fields (SDFs). Historical approaches involve lightfields~\cite{szeliski1996lumigraph,levoy1996lightfield} and coloured voxel grids~\cite{szeliski1998stereo,seitz1999photorealistic}. Implicit neural representations leveraging neural networks to encode the scene~\cite{Mescheder2019occupancy,Park2019deepsdf}.
\textbf{Neural Radiance Fields (NeRFs)} are a recent popular advancement due to their ability for high-fidelity view synthesis from RGB~\cite{mildenhall2021nerf,muller2022instant,barron2022mip} or RGBD~\cite{roessle2022dense,deng2022depth,jung2023importance} images.
Accurate camera poses are crucial to train NeRFs. Typical setups extract poses using structure-from-motion~\cite{schoenberger2016sfm} which is limited in homogeneous regions and to visual ambiguities~\cite{manhardt2019explaining}. Relaxed methods like BARF~\cite{lin2021barf}, Nerf--~\cite{wang2021nerf}, or dynamic SLAM~\cite{karaoglu2023dynamon} can help to robustify camera pose retrieval.
In line with existing literature~\cite{park2021nerfies,li2022neural,weng2022humannerf,chen2023uv}, we adopt the convention of assuming either static cameras or posed images.

In comparison, \textbf{3D Gaussian splatting} has emerged very recently as a compelling \textbf{explicit method} to represent 3D scenes~\cite{kerbl20233d}. This technique capitalizes on the co-produced COLMAP point cloud to represent the scene using 3D Gaussians, optimizing their anisotropic covariance and incorporating visibility-aware rendering.

\subsection{Dynamic Scenes}
Dynamic scenes are also considered both to render novel camera views of humans~\cite{carranza2003free} and their reconstruction~\cite{starck2007surface}. A series of works by researchers from Microsoft~\cite{collet2015high,dou2016fusion4d,orts2016holoportation} introduces content-agnostic streamable free-viewpoint systems with multi-view RGBD cameras.

Also NeRFs have been extended to the temporal domain from synchronized \textbf{multi-view} video~\cite{li2022neural,pumarola2021d}.
Nerfies~\cite{park2021nerfies} addresses the challenging case of a \textbf{moving monocular camera} with limited scene dynamics as video input and optimizes a 5D NeRF. HyperNeRF~\cite{park2021hypernerf} offers a higher-dimensional representation that also accounts for topological changes.
Efficient 4D NeRF rendering methods allow interactive frame rates~\cite{Song2023nerfplayer} with high resolution~\cite{xu20234k4d}.
3D Gaussian splatting has also been extended in time~\cite{wu20234d,yang2023deformable3dgs} by representing generic 4D scenes and leveraging temporal connections. Additionally, 3D Gaussians can be densely tracked over time~\cite{luiten2023dynamic}. In our work, we take a different approach, utilizing a deformation field in conjunction with a parametric human model. This allows us to associate dynamic 3D Gaussians in a canonical representation, enabling the high-fidelity learning of 3D avatars that can be freely animated using an underlying parametric 3D human model.

\subsection{Humans in 3D}
The representation of humans in 3D space has seen significant advancements through the development and extension of spatial human body models in the past decade. Early methods focused on predicting major body joints~\cite{insafutdinov2016deepercut} and keypoints including hand, face, and foot to retrieve human skeleton motion~\cite{openpose}. With SMPL~\cite{loper2015smpl} a widely adopted \textbf{parametric shape representation} has been introduced, utilizing skinning techniques and blend shapes optimized on real 3D body scans, providing realistic representations for non-clothed humans.

These advancements have been paralleled by substantial efforts in dataset acquisition, leveraging advanced hardware setups, accurate annotations, and synchronized multi-view capture. Early work by Kanade et al.~\cite{kanade1997virtualized} utilized a multi-camera dome for reconstruction. Subsequent datasets like Human3.6M~\cite{ionescu2013human3p6}, ZJU-Mocap~\cite{peng2021neural}, THUman4.0~\cite{zheng2022structured}, and ActorsHQ~\cite{icsik2023humanrf} build on these experiences and provide human captures with multi-view cameras.
However, it is important to note that even with advanced hardware, obtaining high-quality 3D annotations remains a costly endeavor~\cite{wang2022phocal,jung2022housecat6d,icsik2023humanrf}.

In the context of registering 3D shapes, \textbf{scene-agnostic} methods typically rely on point cloud descriptors~\cite{drost2010model,rusu20113d,salti2014shot,guo2020deep} that can be used for both 3D and 4D descriptor matching~\cite{deng2018ppfnet,yu2021cofinet,qin2022geometric,saleh2022bending,yu2023rotation}, or on functional maps applied to known shapes~\cite{litany2017deep,bastian2023s3m}. Establishing the connection to mentioned parametric models and finding correspondences to canonical parts plays a crucial role in enabling subsequent 3D avatar animation.

\subsection{Animatable Human Avatars}
The pursuit of animatable human avatars has seen a progression from static 3D human surface recovery to more intricate neural approaches guided by 3D human models.
Deepcap~\cite{habermann2020deepcap} focuses on deformable pre-scanned humans for monocular human performance capture, leveraging weak multi-view supervision, while Monoperfcap~\cite{xu2018monoperfcap} provided the first marker-less method for temporally coherent 3D performance capture of a clothed human from monocular video.
The emergence of neural methods specializing in human rendering marked a significant shift. Models like Neural Body~\cite{pumarola2021d} optimize latent features for each mesh vertex, leveraging an SMPL model. Approaches, including Neural Articulated Radiance Field~\cite{noguchi2021neural}, A-NeRF~\cite{su2021nerf}, and TAVA~\cite{li2022tava} provide solutions to learn human shape, appearance, and pose using skeleton input.
PoseVocab~\cite{li2023posevocab} further learns an optimal pose embedding for dynamic human motion from multi-view RGB by factorizing global pose into embeddings for each joint together with a strategy to interpolate features for fine-grained human appearance.

For monocular video, Selfrecon~\cite{jiang2022selfrecon} combines explicit and implicit representations using SDFs for 3D human body representations, while NeuMan~\cite{jiang2022neuman} splits the scene content to train two NeRFs one for the static background and one for the human. Animatable NeRF~\cite{peng2021animatable} and HumanNerf~\cite{weng2022humannerf} integrates motion priors in the form of SMPL~\cite{loper2015smpl} parameters to regularize field prediction.
The canonical representation is warped into the target frame by Arah~\cite{wang2022arah}, SNARF~\cite{chen2021snarf} as well as Scanimate~\cite{saito2021scanimate} while others~\cite{liu2021neural,xu2022surface} use backward sampling from the observation.
MonoHuman~\cite{Yu_2023_CVPR} models deformation bi-directionally to retrieve view-consistent avatars.
Typically garment deformation is tied to the model, however, Zheng et al.~\cite{zheng2022structured} use a residual model correction to fit dynamic cloth deformations with local fields attached to pre-defined SMPL nodes.
Inherent texture fidelity limitations have been additionally addressed with generative models~\cite{wang2021towards,svitov2023dinar}. All these approaches prioritize rendering quality but also grapple with high computational costs for both training and testing. 
NeRFs are great to generate texture on the fixed geometry~\cite{chen2023texpose}. UV Volumes~\cite{chen2023uv} leverages this to establish a rendering pipeline for human avatars by generating texture and its mapping separately.
The popularity of implicit neural fields for scene representations undoubtedly lead to impressive results. However, these methods inherently struggle to represent empty space efficiently and acceleration is limited by efficient sampling for ray-marching and the structured representations of the scene.
Using explicit 3D Gaussians allows us to overcome these bottlenecks with an unstructured representation for more efficient training with higher quality and faster inference speed.

Concurrent to \OURS, D3GA~\cite{zielonka2023drivable} has just proposed to compose 3D Gaussians embedded in tetrahedral cages for cloth layer decomposition of human avatars. They use multi-view video captures from 200 synchronized cameras as input while \OURS~works with as little as one monocular view.

\begin{figure*}[!t]
 \centering
    \includegraphics[width=1.0\linewidth]{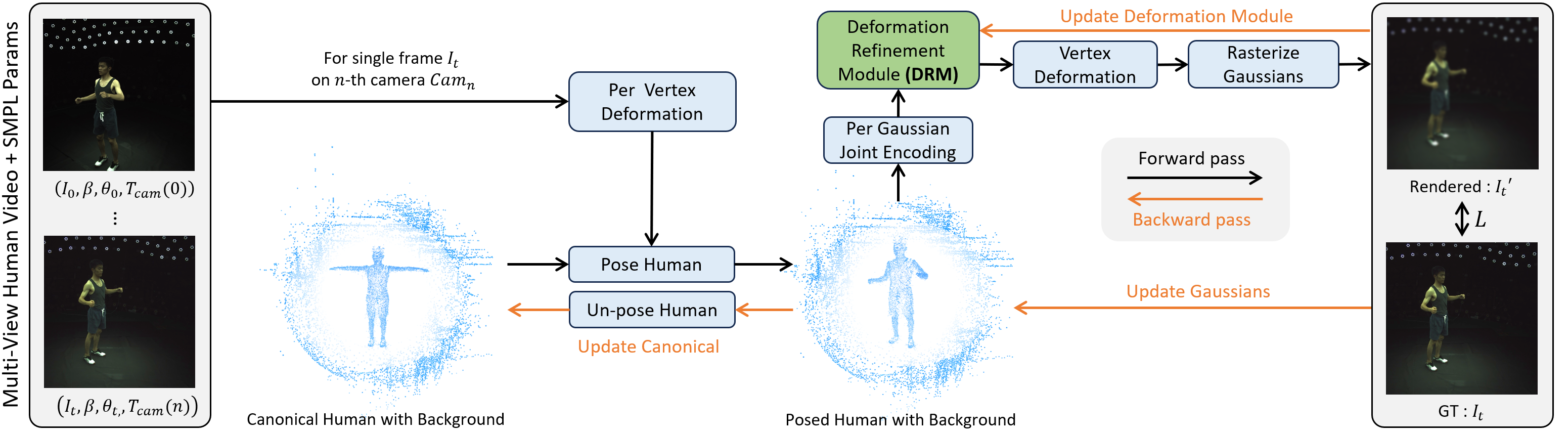}
    \vspace*{-5mm}
    \caption{\textbf{Training Pipeline Overview.} \OURS~is a fully explicit animatable human representation based on 3D Gaussians~\cite{kerbl20233d}. Information from images $I_j$, $1\leq j \leq t$ of the $n$-th camera $\text{Cam}_n$ are integrated into the avatar by using the camera pose $T_{Cam}(j)$, human shape $\beta$, and pose $\theta_{j}$ parameters (left).
    Correspondences between Gaussians of a Canonical and Posed Human are established by a Per Vertex Deformation Module (centre to left, black arrows).
    Residual corrections of Gaussians are performed using a Deformation Refinement Module (DRM) (centre to right, black arrows) before image synthesis through rasterization (right).
    The rendered output can then be compared to ground truth input images to calculate gradients and update both the DRM and human avatar (orange arrows)}
    \label{fig:pipeline}
\end{figure*}

\section{Method}
 
\label{sec:method}

To achieve lifelike animatable human avatars, we need to map low-frequency features such as human pose and shape to high-frequency appearance details.
\OURS~leverages 3D Gaussian Spatting~\cite{kerbl20233d} (3D-GS) as the foundational framework for three primary reasons. \textbf{1.} 3D Gaussians, initialized from coarse point clouds can be associated to vertices of a parametric human model. \textbf{2.} The point cloud structure of Gaussians facilitates straightforward deformation based on human parameters. \textbf{3.} 3D Gaussians enables swift and efficient rendering of high-frequency details. Our pipeline overview is depicted in Fig.~\ref{fig:pipeline}.

\subsection{3D Gaussian Initialization} \label{subsec:gaussian_recap_init}

3D Gaussians are an explicit scene representation that contains the following attributes: geometric center $\mathbf{P}_{i} \in \mathbb{R}^3$, rotation $\mathbf{R}_{i} \in \text{SO}\left(3\right)$, size, scale, opacity, spherical harmonics (SH).
Images can be rendered from this representation with a point-based rendering scheme.
While being similar to surfels~\cite{sufel}, Gaussians incorporate a 3-dimensional scale vector to create elliptical shapes of different size and shape and SH can model view dependent color variations. Originally, Gaussians are first initialized with SfM~\cite{schoenberger2016mvs, schoenberger2016sfm} point clouds and then trained with a specific adaptive density control scheme that splits, clones and prunes the Gaussians during training depending on their size and gradient magnitude.
This takes an important role in optimizing the number of Gaussians that is denser in detailed regions and coarser in larger homogeneous regions.

Unlike the original implementation~\cite{kerbl20233d}, we use two different point clouds (Fig.~\ref{fig:gaussian_overview} (a)) to initialize Gaussians and include additional features: \textit{parent index $i$} and \textit{surface normal $\mathbf{n}_i$} (only for human).
For the background, we initialize random points on a spherical surface and label parents with $i = b$.

For the human, we generate a parametric human mesh in canonical T-pose \(M^{\text{SMPL}}_{\theta_{c}}\) using SMPL~\cite{loper2015smpl}, then initialize the canonical Gaussians \(G_c\) at the center of each mesh face and assign extra features such as the face normal and parent as an index of the face in the mesh.

\subsection{Posing Gaussians} \label{subsec:posing_gaussian}
After initialization, each Gaussian is deformed according to its parent using Per Vertex Deformation (PVD). Given the \mbox{\(t\)-th} human pose parameter \(\theta_{t}\), a new human mesh \(M^{\text{SMPL}}_{\theta_{t}}\) is generated from SMPL~\cite{loper2015smpl} and per face human deformation $D^{i}_{t}$ is calculated for all faces $i \in \mathcal{F}$ of the SMPL model by comparing with the canonical human face:
\begin{equation}
    \mathbf{D}_{t} =
    \left( D^{1}_{t}, \ldots, D^{F}_{t} \right) =
    %\left( q, t \right)^{f} =
    \text{PVD} \left( M^{\text{SMPL}}_{\theta_{c}},M^{\text{SMPL}}_{\theta_{t}}\right).
\end{equation}
The deformed Gaussians \(G_{d}\) can then be obtained by transforming the canonical Gaussians \(G_c\) into their new locations according to $\mathbf{D}_{t}$.
Associated locations $\mathbf{P}_{i} \in \mathbb{R}^3$ and rotations $\mathbf{R}_{i} \in \text{SO}\left(3\right)$ can be established through $D^{i}_{t}$ via:
\begin{align*}
    G_{d}(\mathbf{P}_{i},\mathbf{R}_{i})
    = D^{i}_{t} \left( G_{c} \left( \mathbf{P}_{i},\mathbf{R}_{i} \right) \right).
\end{align*}
\begin{figure*}[!t]
 \centering
    \includegraphics[width=1.0\linewidth]{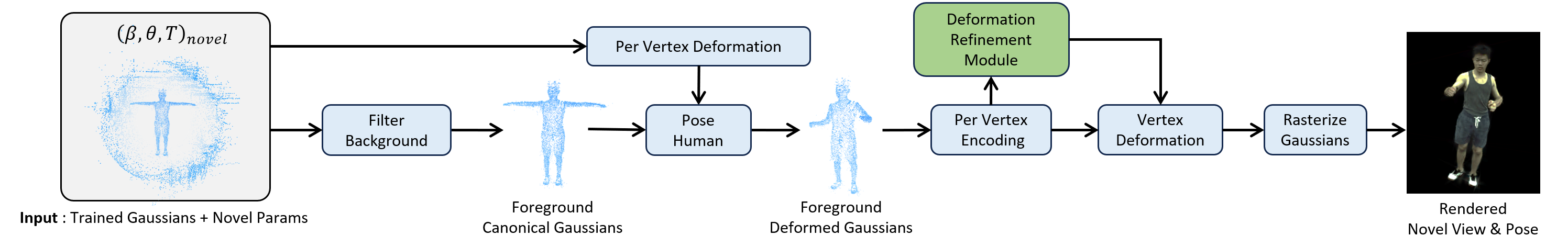}
    \vspace*{-6mm}
    \caption{\textbf{Inference Pipeline Overview.} During inference time, we first filter out the background Gaussians (left) and then deform the canonical human avatar. A coarse deformation is done first using SMPL~\cite{loper2015smpl} parameters followed by the DRM correction (centre to right). The output is an animated human without background (right).}
    \label{fig:inference_pipeline}
\end{figure*}
\subsection{Deformation Refinement}\label{subsec:deformation_refine}
For humans with tight clothing, it is enough to deform with an SMPL~\cite{loper2015smpl} model. However, in general garment motion causes more deformations dependent on the human posture.
To obtain higher fidelity renderings, we include a Deformation Refinement Module (DRM).
The module is designed with a small set of MLPs that take the distance between the center of the Gaussian $\mathbf{P}_{i}$ and the human joints $\mathbf{J}_{t}$ of the SMPL model to predict a residual refinements for PVD deformations.
In practice, using shorthand notation $G_{d}(\mathbf{P}_{i})$ for the location of the Gaussians after the deformation, the feature
\begin{equation}\label{eq:encoding}
    \mathbf{E}^{i}_{d} =  G_{d}(\mathbf{P}_{i}) - \mathbf{J}_{t}
\end{equation}
is flattened and fed into an MLP to calculate the corrected per Gaussian transformation $\mathbf{D}_{r} = \left( D^{1}_{r}, \ldots, D^{F}_{r} \right)$ with
\begin{equation}
    D^{i}_{r} = \text{DRM} \left( \mathbf{E}^{i}_{d} \right).
\end{equation}
The detailed architecture is specified in the supplementary material.
Finally $D^{i}_{r}$ is applied to $G_{d}$ to obtain the refined Gaussians $G_{r}$ as
\begin{equation}
    G_{r}(\mathbf{P}_{i},\mathbf{R}_{i}) =  D^{i}_{r} \left( G_{d} \left( \mathbf{P}_{i},\mathbf{R}_{i} \right) \right). 
\end{equation}
This refinement step plays a crucial role in producing high fidelity renderings of clothed human avatars.
Fig.~\ref{fig:ablation_drm} shows the impact of this residual correction step.

\subsection{Spherical Harmonics Direction} 
In 3D-GS~\cite{kerbl20233d}, Spherical Harmonics (SH) are used to incorporate viewing angle dependent effects. The direction $\mathbf{d}_i \in \mathbb{R}^3$ is calculated as the relative vector from the camera center $\mathbf{c} \in \mathbb{R}^3$ to each Gaussian via
\begin{equation}\label{eq:dir_rel}
    \mathbf{d}^{i} =  \frac{\mathbf{c} - \mathbf{P}_{i}}{\left\lVert \mathbf{c} - \mathbf{P}_{i} \right\rVert_2}.
\end{equation}
While this assumption holds for static scenes, it fails in dynamic scenarios where the same point can have the identical relative direction to a camera but might be viewed from a different angle due to surface rotations.
We correct this by including surface normal information calculated from the SMPL~\cite{loper2015smpl} prior mesh (Sec.~\ref{subsec:gaussian_recap_init}).
We apply the rotations $\mathbf{R}_{i}^{t}$, $\mathbf{R}_{i}^{r}$ from the deformations $\mathbf{D}_{t}$, $\mathbf{D}_{r}$ sequentially on the surface normal $\mathbf{n}_i$ to obtain more realistic corrected directions $\mathbf{d}^{i}_c \in \mathbb{R}^3$ by
\begin{equation}\label{eq:dir_normal}
    \mathbf{d}^{i}_c = \mathbf{R}_{i}^{r} \cdot \mathbf{R}_{i}^{t} \cdot \mathbf{n}_i.
\end{equation}
We use both direction definitions for rasterization.
Eq.~\ref{eq:dir_rel} is used on the static background while Eq.~\ref{eq:dir_normal} is used for the dynamic foreground.
\subsection{Un-Posing Gaussians and Updating Parents}
Once the Gaussians are updated, they are transformed back to the canonical space in order to be posed again with the next set of parameters.
Un-posing is done by following the deformation in the inverse order.
From the updated Gaussians, we apply the inverse of $\mathbf{D}_{t}$ and $\mathbf{D}_{r}$ sequentially to transform \(G_{r}\) into the updated canonical Gaussians \(G_{c}\).

During training, the number of Gaussians and their centers may change and the parent index $i$ has to be updated accordingly.
When Gaussians are added via splitting and cloning~\cite{kerbl20233d}, the parent index is passed to the new Gaussians.
Parent assignments are updated every 1000-th iteration by finding the face with the shortest Euclidean distance in the canonical pose.
This prevents Gaussians from diverging too far from their parents during the update of $\mathbf{P}_{i}$.
Gaussians with a distance above a threshold $\tau$ from the surface are assigned to the background.

\subsection{Loss Design and Inference Pipeline} \label{subsec:loss}
Three losses are applied throughout the training: L1 loss \(\mathcal{L}_{1}\), structural similarity loss \(\mathcal{L}_{\text{SSIM}}\), and perceptual similarity loss $\mathcal{L}_{\text{LPIPS}}$ with AlexNet~\cite{Krizhevsky2012ImageNetCW} as the base.
The losses are balanced by the weights $\lambda_{\text{L1}}$, $\lambda_{\text{SSIM}}$, and $\lambda_{\text{LPIPS}}$.

The overall training objective reads as follows:
\begin{equation}\label{eq:loss}
    \mathcal{L}
    = \lambda_{\text{L1}}\mathcal{L}_{1}
    +\lambda_{\text{SSIM}}\mathcal{L}_{\text{SSIM}}
    +\lambda_{\text{LPIPS}}\mathcal{L}_{\text{LPIPS}}.
\end{equation}
 During inference time (cf.~Fig.~\ref{fig:inference_pipeline}), our pipeline filters the background Gaussians based on their parents. Taking a novel camera pose and human pose as input the Gaussians are deformed based on the SMPL model first and then adjusted via the DRM. The deformed Gaussians are then passed to a rasterizer to render the human under a given camera pose.
\section{Training and Implementation Details} \label{sec:training}
For training, we calculate the loss by comparing the rendered image with the ground truth image and then update the Gaussian features~\cite{kerbl20233d} as well as the DRM based on the resulting gradients. As both updates involve displacements of Gaussians, we split the training into multiple stages and mutually update these modules.

\subsection{Training Schedule} \label{subsec:train_schedule}
We begin by updating both DRM and Gaussians at the same time until 10k-iterations to initialize them. After that, we mutually fix either the Gaussians or DRM and update its counterpart. We switch the weight updates between them every 5k-iterations until 100k-iterations. During the Gaussian part, alongside the Gaussian features update, we follow the original Gaussian adaptive update schemes~\cite{kerbl20233d} until 50k iterations.

\section{Experiments}
\label{sec:experiments}
We first describe the experimental setup including datasets and baseline methods, then show qualitative and quantitative results as well as train and test efficiency comparisons.

For all our experiments, we use the parameters $\tau = 10~\text{cm}$ as well as weights \(\lambda_{L1}=0.6\), \(\lambda_{SSIM}=0.4\), and \(\lambda_{LPIPS}=0.4\).

\subsection{Data Selection and Evaluation Setup}
To illustrate the background handling of \OURS, we choose two human datasets with different background setups (ZJU-MoCap~\cite{peng2021neural} and THUman4.0~\cite{zheng2022structured}).
The \textbf{ZJU-MoCap}~\cite{peng2021neural} dataset features a motion capture sphere setup with 21-23 cameras with 9 human subjects performing actions of different length (600-2500 frames) in front of a relatively dark background.
The data helps to understand how our method can differentiate the human from the background in a low light scenario where human subjects often blend into the dark background.
The data of \textbf{THUman4.0}~\cite{zheng2022structured} features 24 cameras with only 3 human subjects performing relatively long sequences of actions (2500-5060 frames) in an uncontrolled and partially unbounded indoor scene. We specifically select this dataset to evaluate our performance in less constrained, more realistic scenes.

\begin{figure}[!hh]
 \centering
    \includegraphics[width=1.0\linewidth]{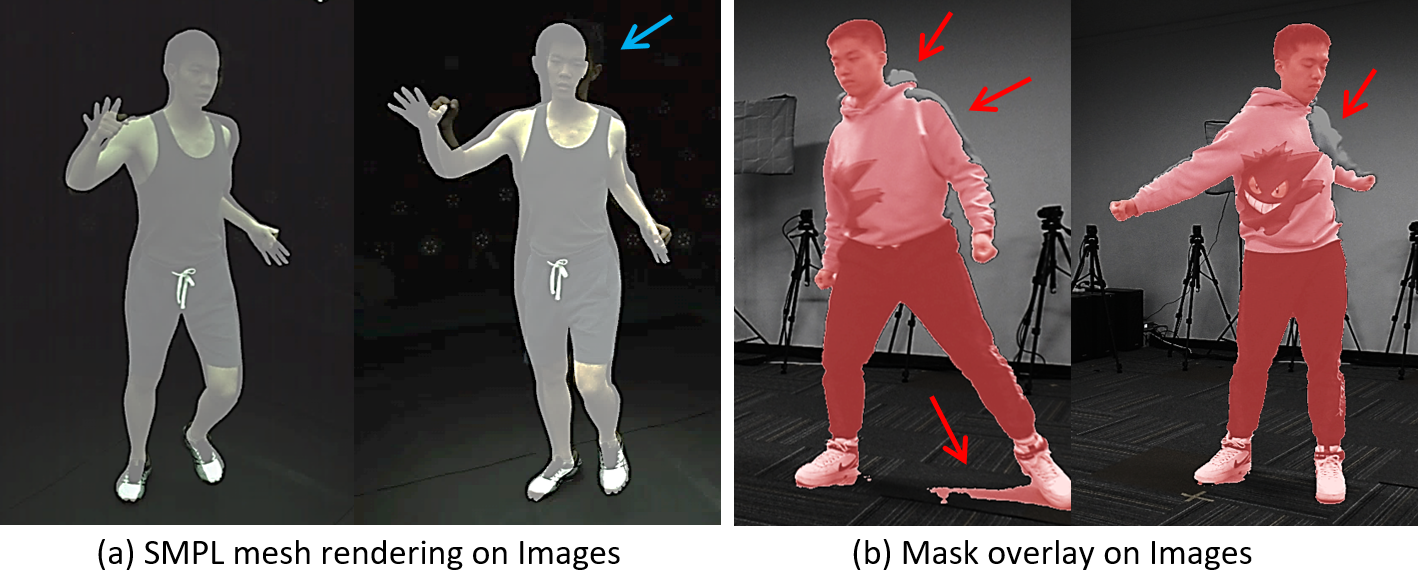}
    \vspace*{-6mm}
    \caption{\textbf{Issues in Datasets.} Some scenes in the ZJU dataset~\cite{peng2021neural} suffer from inaccurate extrinsic calibration. The rendered SMPL meshes on these images are not consistent over cameras (a). On the other hand, the mask in the THUman dataset~\cite{zheng2022structured} suffers from over- and under-segmentation artifacts depending on the lighting conditions (b).}
    \label{fig:camera_issue}
\end{figure}

To analyse the performance with only a few views and sparse image sampling, we train the method using seven views and use every 50-th frame per camera as training input.
We further analyse unseen human poses that are significantly different from the training poses (cf. Sec.~\ref{subsec:evaluation}).
Since the calibration quality of the camera views in the ZJU dataset~\cite{peng2021neural} varies severely (cf. Fig~\ref{fig:camera_issue}), we select the best seven views upon visual inspection.
We equally take seven cameras for THUman4.0~\cite{zheng2022structured} to have a consistent number of training views. We specify the cameras that are used for training and testing in the supplementary material.

\subsection{Baseline Selections}
Two of the most recent state-of-the-art methods are selected as baselines: UV-Volumes~\cite{chen2023uv} and PoseVocab~\cite{li2023posevocab}.
Although both methods tackle dynamic human reconstruction and reposing using implicit volume rendering technique, they are based on orthogonal approaches.

\noindent\textbf{UV-Volumes}~\cite{chen2023uv} encodes SMPL~\cite{loper2015smpl} vertices directly into a volume using a sparse 3D CNN. This approach breaks down the human rendering task into two smaller tasks, volume and texture generation. The volume generator conditioned on SMPL~\cite{loper2015smpl} vertices first generates a UV volume via a sparse 3D CNN that renders a UV map from a given camera pose. A consecutive texture generator predicts the texture with SMPL~\cite{loper2015smpl} parameters and body parts mapped to the image using a UV map. Training requires a UV map and body part segmentation as predictions of DensePose~\cite{guler2018densepose}.

\noindent\textbf{PoseVocab}~\cite{li2023posevocab} in comparison encodes features in a canonical pose and samples corresponding features when a human pose is given. Given human poses with joint angles, a pose vocabulary (PoseVocab) in canonical human pose is constructed and takes pose embeddings as queries. When a query pose vector is given, the pose is converted into an embedding and features are sampled from the vocabulary. A volume rendering module synthesizes a human image under the corresponding pose. PoseVocap requires depth maps for full training, which forces the training to be split into two parts. Depth prediction is learned before the full network is trained.

\begin{figure}[!hh]
 \centering
    \includegraphics[width=1.0\linewidth]{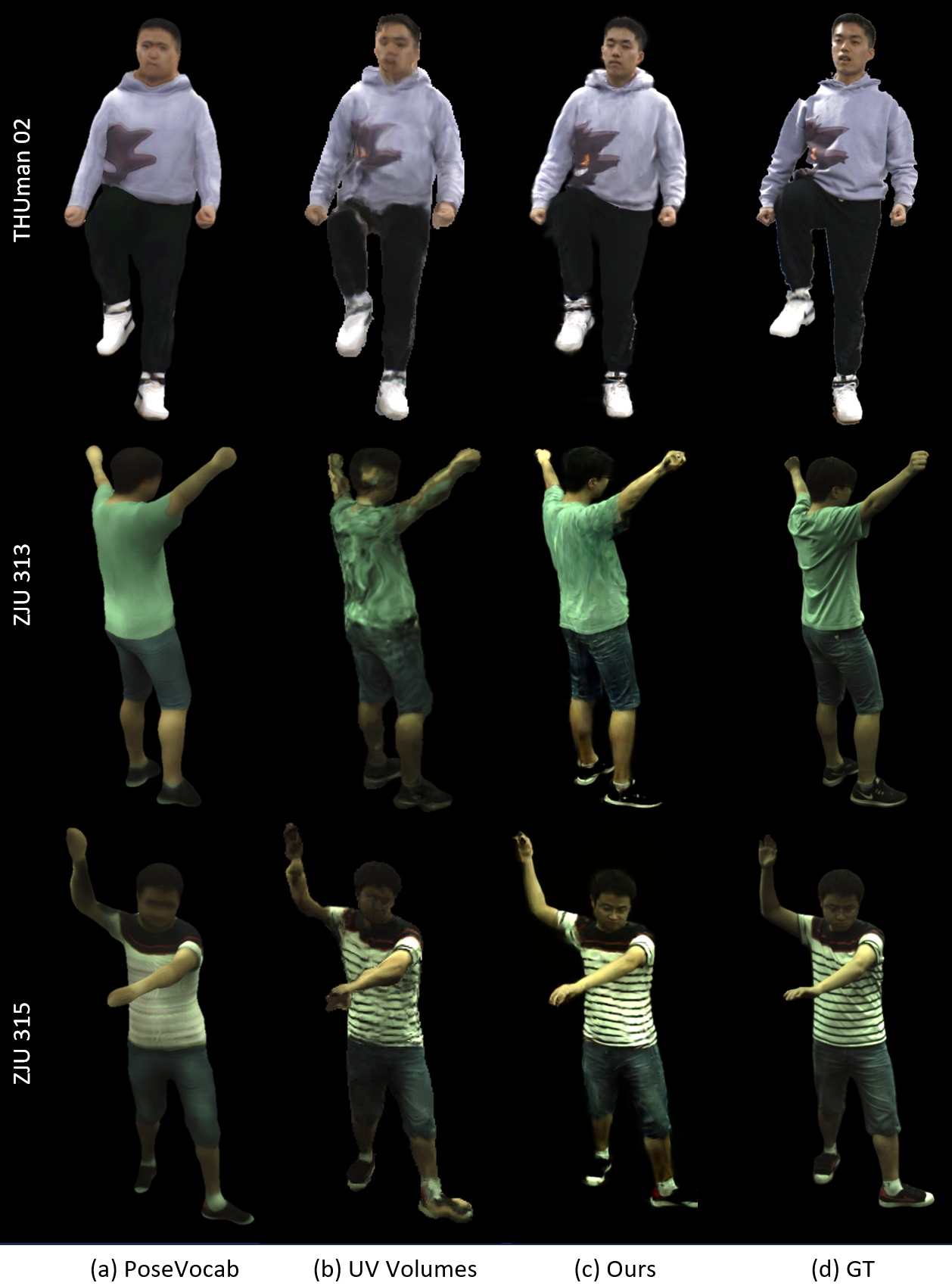}
    \vspace*{-5mm}
    \caption{\textbf{Qualitative Evaluation in a spare view training setting.} PoseVocab~\cite{li2023posevocab} looses details if provided with a sparse setup. UV-Volumes~\cite{chen2023uv} produces more realistic shading by leveraging a texture map, however, suffers from geometric and texture artifacts especially if the inferred human pose is far from the training poses. Ours is limited on generating exact shading (ZJU~\cite{peng2021neural} 313, 315), but produces more realistic and detailed humans in novel poses.} 
    \label{fig:qualitative_evaluation}
\end{figure}

\subsection{Quantitative and Qualitative Evaluation}\label{subsec:evaluation}
For a quantitative evaluation, we follow the literature and use three metrics: Signal to noise ratio (\textbf{PSNR}, higher is better), Structural Similarity (\textbf{SSIM}, higher is better), Perceptional Loss (\textbf{LPIPS}, lower is better).
For the evaluation, we use the furthest in between frames from the training sequence (i.e. every (50\(n\) + 25)-th frame) and evaluate with a mixture of used and novel camera views. See supplementary material for more details.

\noindent\textbf{Ours vs Baselines.} The result in Tab.~\ref{tab:qunti} shows that \OURS~gives much better LPIPS score and similar or better SSIM compared to the baselines indicating that novel pose synthesis is visually perceived as a more human like and maintains good details.
The difference is clearly visible in Fig.~\ref{fig:qualitative_evaluation}.
The PSNR metric on the ZJU dataset~\cite{peng2021neural} for \OURS, however, is lower on average compared to UV-Volumes~\cite{chen2023uv}. The ZJU dataset is recorded in dim lights that leads to a varying shade effect depending on the orientation and pose of the human performer. In this case, UV-Volumes~\cite{chen2023uv}'s dedicated texture generation network has an advantage. It remembers the orientation and pose dependent shading which are reproduced at inference time while the shading in our pipeline does not vary severely.

ZJU 313, 315 in Fig.~\ref{fig:qualitative_evaluation} indicate that our method produces an average shading that can be brighter than GT image.
While UV-Volumes~\cite{chen2023uv}'s replication of a similar shading leads to better PSNR values, it fails to correctly synthesize the human shape.
In comparison, when training in bright lighting (i.e. THUman~\cite{zheng2022structured}, 00-02 in Tab.~\ref{tab:qunti}), \OURS~outperforms the PSNR values of UV-Volumes~\cite{chen2023uv} by a large margin.

\noindent\textbf{Baseline Comparision.} UV-Volumes~\cite{chen2023uv} directly encodes feature volumes using prior SMPL~\cite{loper2015smpl} vertices.
It specializes to replicate seen human poses while it struggles with larger pose deviations. PoseVocab~\cite{li2023posevocab} encodes features in a canonical volume which generalize better to novel human poses, but requires many human poses to train. This can be observed in Tab.~\ref{tab:qunti} and Fig.~\ref{fig:qualitative_evaluation}. UV-Volume replicates shading easily (PSNR\(\uparrow\)) while the human shape breaks (LPIPS\(\uparrow\)). PoseVocap's rendering lack details (PSNR\(\downarrow\)) due to the limited training images while the overall human shape is preserved (LPIPS\(\downarrow\)).

\noindent\textbf{Rendering Efficiency} For all tests, we use a consumer grade laptop (i9 + RTX3080 max-Q) and measure the inference time. \OURS~and PoseVocab~\cite{li2023posevocab} were capable of rendering at full resolution. \OURS~takes roughly 0.3~sec while PoseVocab requires 15-25~sec per image, which is a speed up of $>50\times$. Due to the high GPU memory requirement of UV-Volumes, we were not able to render at full resolution. At 30\% resolution, the method shows a similar inference time compared to our approach.

\begin{table}[]
\caption{\textbf{Quantitative Evaluation for Novel Pose Synthesis.} We compare our method against two recent state of the art works, UV-Volumes~\cite{chen2023uv} (UVV) and PoseVocab~\cite{li2023posevocab} (PV), on ZJU-Mocap~\cite{peng2021neural} (row 1-4) and THUman4.0~\cite{zheng2022structured} (row 5-7) dataset.}
\label{tab:qunti}
\resizebox{\columnwidth}{!}{
\begin{tabular}{c|ccc|ccc|ccc}
\hline
\multirow{2}{*}{Scene} & \multicolumn{3}{c|}{PSNR \(\uparrow\)} & \multicolumn{3}{c|}{SSIM \(\uparrow\)} & \multicolumn{3}{c}{LPIPS \(\downarrow\)} \\ \cline{2-10} 
                       & UVV                & PV       & Ours              & UVV       & PV                & Ours              & UVV          & PV           & Ours                  \\ \hline
313                    & \textbf{29.18}     & 22.40    & 28.33             & 0.960     & \textbf{0.961}    & 0.960             & 0.056        & 0.048        & \textbf{0.043}        \\
315                    & \textbf{26.49}     & 21.96    & 24.51             & \textbf{0.957}     & 0.954             & 0.955    & 0.043        & 0.044        & \textbf{0.034}        \\
377                    & \textbf{29.32}     & 23.30    & 28.44             & 0.969     & 0.970             & \textbf{0.974}    & 0.047        & 0.035        & \textbf{0.033}        \\
394                    & 29.12              & 22.74    & \textbf{29.28}    & 0.957     & 0.962             & \textbf{0.964}    & 0.059        & 0.047        & \textbf{0.040}        \\ \hline
00                     & 28.33              & 26.30    & \textbf{30.32}    & 0.961     & 0.970             & \textbf{0.975}    & 0.048        & 0.038        & \textbf{0.028}        \\
01                     & 25.88              & 25.07    & \textbf{28.65}    & 0.944     & 0.962             & \textbf{0.963}    & 0.064        & 0.061        & \textbf{0.047}        \\
02                     & 24.49              & 23.97    & \textbf{26.54}    & 0.942     & 0.959             & \textbf{0.961}    & 0.080        & 0.062        & \textbf{0.057}        \\ \hline
\end{tabular}
}
\end{table}

\subsection{Ablation Study}

\begin{figure*}[!t]
 \centering
    \includegraphics[width=1.0\linewidth]{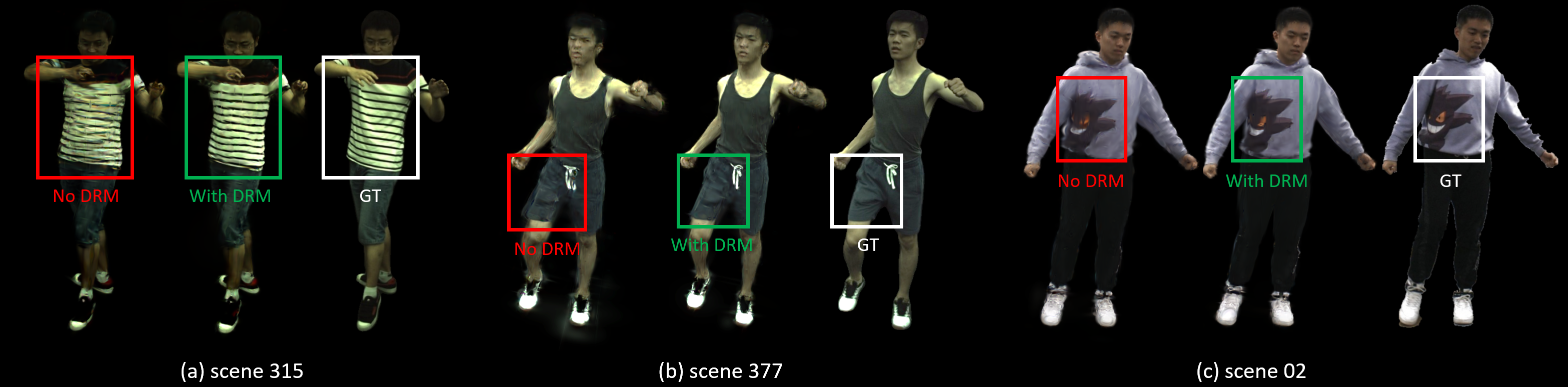}
    \vspace*{-5mm}
    \caption{\textbf{Effect of the Deformation Refinement Module.} The residual correction of the DRM plays an important role in our pipeline to cope with garment deformations. Training without it (red) results in loss of texture details (a). DRM training (green) reduces ghosting effects (b) and improves texture boundaries (c). The ground truth is illustrated in white.}
    \label{fig:ablation_drm}
\end{figure*}

\begin{figure}[!t]
 \centering
    \includegraphics[width=1.0\linewidth]{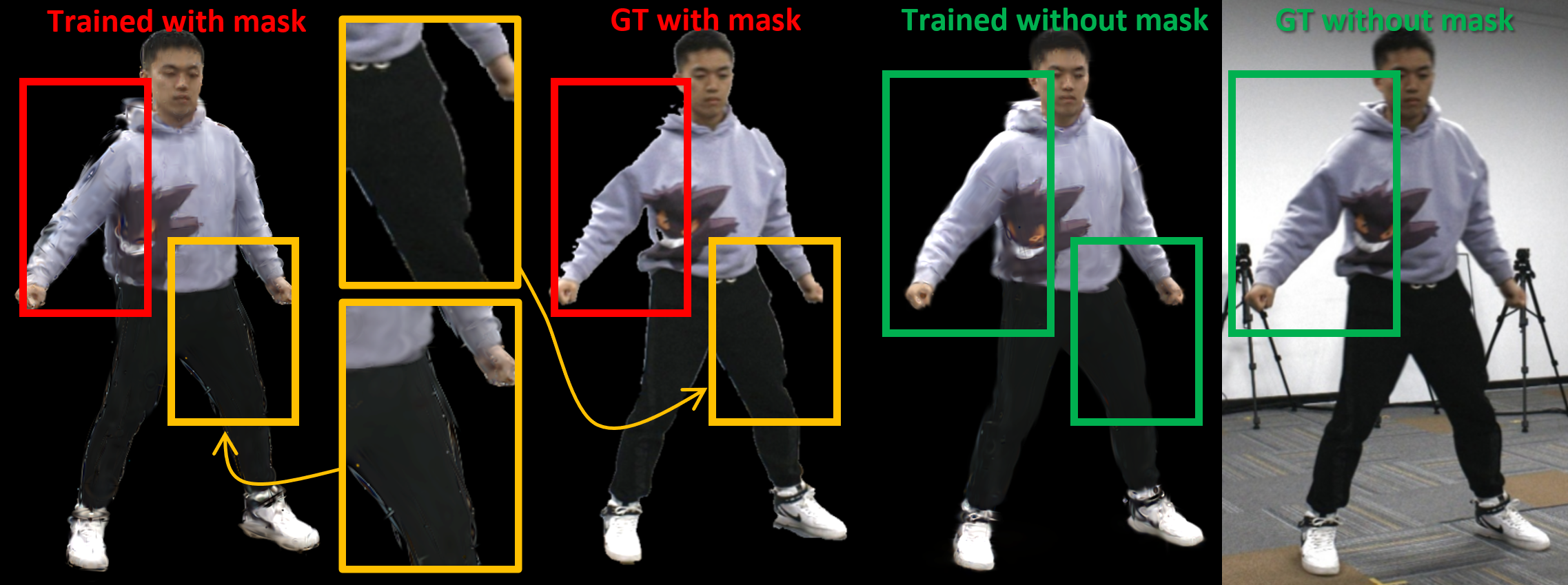}
    \vspace*{-5mm}
    \caption{\textbf{Training with Mask Annotations.} When trained with human masks (left), mask errors (second from left) manifest in black (red box) and white (orange box) Gaussians around the body. If \OURS~is trained without these annotations (second from right), it produces higher quality geometry of the human shape.}
    \label{fig:ablation_mask}
\end{figure}

\begin{table}[!t]
\caption{\textbf{Ablation Study.} We run two ablations for DRM and masks. Ours wo.D denotes \OURS~trained without DRM while w.M indicates training with mask. In general, our full pipeline (Full) does not perform best quantitatively under these metrics, however, gives clearly better visual results (Fig.~\ref{fig:ablation_drm},~\ref{fig:ablation_mask}).}
\label{tab:ablation}
\resizebox{\columnwidth}{!}{
\begin{tabular}{c|ccc|ccc|ccc}
\hline
\multirow{2}{*}{Scene} & \multicolumn{3}{c|}{PSNR \(\uparrow\)} & \multicolumn{3}{c|}{SSIM \(\uparrow\)} & \multicolumn{3}{c}{LPIPS \(\downarrow\)} \\ \cline{2-10} 
                       & Full      & wo.D              & w.M               & Full            & wo.D           & w.M            & Full            & wo.D            & w.M             \\ \hline
315                    & 24.54     & 25.45             & \textbf{25.82}    & 0.955           & 0.953          & \textbf{0.962} & 0.034           & 0.038           & \textbf{0.029}  \\
377                    & 28.44     & \textbf{28.96}    & 27.17             & \textbf{0.984}  & 0.976          & 0.960          & 0.033           & \textbf{0.031}  & 0.058           \\ \hline
02                     & 26.54     & \textbf{26.77}    & 26.67             & 0.961           & \textbf{0.962} & 0.960          & 0.057  & 0.056           & \textbf{0.046}\\ \hline
\end{tabular}}
\end{table}

\noindent\textbf{Residual Correction.} The DRM module plays a crucial role in producing texture details when rendering clothed humans. Canonical Gaussians that only deform with the SMPL model are not flexible enough to deal with garment deformation. If trained without this correction, texture details become blurry (Fig.~\ref{fig:ablation_drm}a,c) and ghosting effects at surface boundaries arise (Fig.~\ref{fig:ablation_drm}b).
This obvious improvement is not well quantified using current metrics (Tab.~\ref{tab:ablation}, Full vs wo.D) despite its clear visual advantage (Fig.~\ref{fig:ablation_drm}, green vs red).

\noindent\textbf{Background} The annotation of masks in dynamic sequences is non-trivial resulting in imperfect masks (Fig.~\ref{fig:camera_issue}). These impact training and evaluation. \OURS~can be trained without masks. We perform an extra experiment utilizing masked images to show the impact of the masks on the final result. Fig.~\ref{fig:ablation_mask} shows that unreliable masks can create black (red box) or white Gaussians (orange box) around the human. This artifact disappears in the mask-free training setup.
Since evaluation is performed on incorrectly masked images of the dataset, this is not quantified in Tab.~\ref{tab:ablation} (Full vs w.M).

\begin{figure}[!t]
 \centering
    \includegraphics[width=1.0\linewidth]{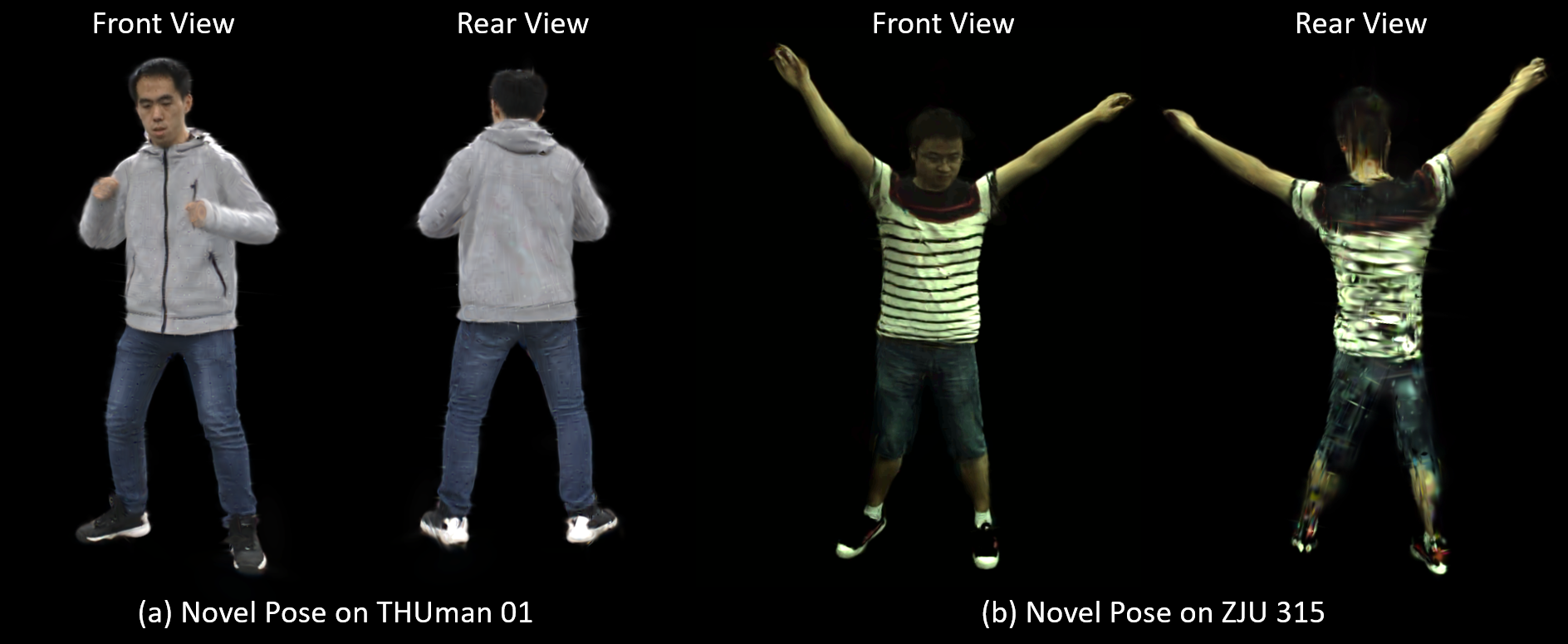}
    \vspace*{-5mm}
    \caption{\textbf{Monocular Training Setup.} Our method can be trained with only a single static camera. Our avatar can be re-posed well in novel views (a) if the individual turns around fully during the training. Unseen visual appearances from areas not seen during training are not recovered (b).}
    \label{fig:ablation_single_view}
\end{figure}

\noindent\textbf{Training with a Single Camera.} We have seen that our method requires significantly fewer views and human poses for training. However, setting up synchronized and calibrated cameras for data capture can be challenging (cf. Fig.~\ref{fig:camera_issue}a). We train our method with only a monocular fixed camera to simulate a real-life recording scenario. To compensate for the 7 times fewer views, we train with every 7th frame of the video sequence. Fig.~\ref{fig:ablation_single_view} shows that our method successfully creates an animatable avatar when the subject covers a 360 degree motion. On the other hand, our explicit method does not hallucinate unseen sides. A full animated sequence can be found in our supplementary video.

\section{Conclusion and Limitations}
\begin{figure}[!t]
 \centering
    \includegraphics[width=1.0\linewidth]{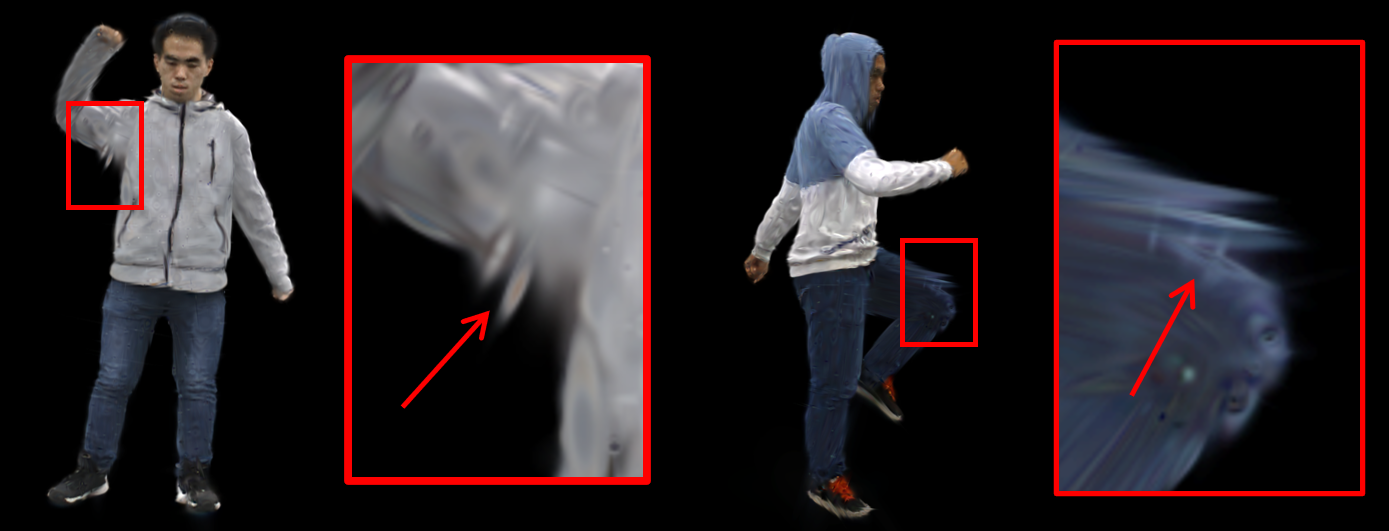}
    \vspace*{-5mm}
    \caption{\textbf{Gaussian Spike Artifact.} Training on uni-coloured garments can cause a reduction of the number of Gaussians while increasing their size. This manifests in Gaussian spikes arising at bent geometry boundaries in unseen poses.}
    \label{fig:limitation}
\end{figure}

In this paper, we propose \OURS, an explicit method for animatable human avatar generation from RGB images. We use deformable 3D Gaussians controlled via SMPL parameters and show that our method can be trained with significantly fewer images, views an without masks compared to prior works.
Even without mask annotations, the method generates high quality 3D digital twins of humans that can be re-posed and rendered from arbitrary viewpoints.
Still, uni-coloured garments with large deformations can pose problems for animations of out-of-distribution poses as shown in Fig.~\ref{fig:limitation}.
Our method efficiently infers full-resolution images even on a consumer laptop which helps democratizing the work, but also poses ethical concerns due to the fact that a single input video is enough to generate a 3D digital human copy. It shares the issue with all other 3D avatar generators that can be abused to create fake videos of real people.
We still firmly believe that the idea of deformable 3D Gaussian fused with a parametric model can inspire other works.
{
    \small
    \bibliographystyle{ieeenat_fullname}
    \bibliography{main}

\begin{thebibliography}{79}
\providecommand{\natexlab}[1]{#1}
\providecommand{\url}[1]{\texttt{#1}}
\expandafter\ifx\csname urlstyle\endcsname\relax
  \providecommand{\doi}[1]{doi: #1}\else
  \providecommand{\doi}{doi: \begingroup \urlstyle{rm}\Url}\fi

\bibitem[Alldieck et~al.(2019)Alldieck, Magnor, Bhatnagar, Theobalt, and Pons-Moll]{alldieck2019learning}
Thiemo Alldieck, Marcus Magnor, Bharat~Lal Bhatnagar, Christian Theobalt, and Gerard Pons-Moll.
\newblock Learning to reconstruct people in clothing from a single rgb camera.
\newblock In \emph{Proceedings of the IEEE/CVF Conference on Computer Vision and Pattern Recognition}, pages 1175--1186, 2019.

\bibitem[Barron et~al.(2022)Barron, Mildenhall, Verbin, Srinivasan, and Hedman]{barron2022mip}
Jonathan~T Barron, Ben Mildenhall, Dor Verbin, Pratul~P Srinivasan, and Peter Hedman.
\newblock Mip-nerf 360: Unbounded anti-aliased neural radiance fields.
\newblock In \emph{Proceedings of the IEEE/CVF Conference on Computer Vision and Pattern Recognition}, pages 5470--5479, 2022.

\bibitem[Bastian et~al.(2023)Bastian, Baumann, Hoppe, B{\"u}rgin, Kim, Saleh, Busam, and Navab]{bastian2023s3m}
Lennart Bastian, Alexander Baumann, Emily Hoppe, Vincent B{\"u}rgin, Ha~Young Kim, Mahdi Saleh, Benjamin Busam, and Nassir Navab.
\newblock S3m: Scalable statistical shape modeling through unsupervised correspondences.
\newblock In \emph{International Conference on Medical Image Computing and Computer-Assisted Intervention}, pages 459--469. Springer, 2023.

\bibitem[{Cao} et~al.(2019){Cao}, {Hidalgo Martinez}, {Simon}, {Wei}, and {Sheikh}]{openpose}
Z. {Cao}, G. {Hidalgo Martinez}, T. {Simon}, S. {Wei}, and Y.~A. {Sheikh}.
\newblock Openpose: Realtime multi-person 2d pose estimation using part affinity fields.
\newblock \emph{IEEE Transactions on Pattern Analysis and Machine Intelligence}, 2019.

\bibitem[Carranza et~al.(2003)Carranza, Theobalt, Magnor, and Seidel]{carranza2003free}
Joel Carranza, Christian Theobalt, Marcus~A Magnor, and Hans-Peter Seidel.
\newblock Free-viewpoint video of human actors.
\newblock \emph{ACM transactions on graphics (TOG)}, 22\penalty0 (3):\penalty0 569--577, 2003.

\bibitem[Chen et~al.(2023{\natexlab{a}})Chen, Manhardt, Navab, and Busam]{chen2023texpose}
Hanzhi Chen, Fabian Manhardt, Nassir Navab, and Benjamin Busam.
\newblock Texpose: Neural texture learning for self-supervised 6d object pose estimation.
\newblock In \emph{Proceedings of the IEEE/CVF Conference on Computer Vision and Pattern Recognition}, pages 4841--4852, 2023{\natexlab{a}}.

\bibitem[Chen et~al.(2021)Chen, Zheng, Black, Hilliges, and Geiger]{chen2021snarf}
Xu Chen, Yufeng Zheng, Michael~J Black, Otmar Hilliges, and Andreas Geiger.
\newblock Snarf: Differentiable forward skinning for animating non-rigid neural implicit shapes.
\newblock In \emph{Proceedings of the IEEE/CVF International Conference on Computer Vision}, pages 11594--11604, 2021.

\bibitem[Chen et~al.(2023{\natexlab{b}})Chen, Wang, Chen, Zhang, Li, Guo, Wang, and Wang]{chen2023uv}
Yue Chen, Xuan Wang, Xingyu Chen, Qi Zhang, Xiaoyu Li, Yu Guo, Jue Wang, and Fei Wang.
\newblock Uv volumes for real-time rendering of editable free-view human performance.
\newblock In \emph{Proceedings of the IEEE/CVF Conference on Computer Vision and Pattern Recognition}, pages 16621--16631, 2023{\natexlab{b}}.

\bibitem[Collet et~al.(2015)Collet, Chuang, Sweeney, Gillett, Evseev, Calabrese, Hoppe, Kirk, and Sullivan]{collet2015high}
Alvaro Collet, Ming Chuang, Pat Sweeney, Don Gillett, Dennis Evseev, David Calabrese, Hugues Hoppe, Adam Kirk, and Steve Sullivan.
\newblock High-quality streamable free-viewpoint video.
\newblock \emph{ACM Transactions on Graphics (ToG)}, 34\penalty0 (4):\penalty0 1--13, 2015.

\bibitem[Deng et~al.(2018)Deng, Birdal, and Ilic]{deng2018ppfnet}
Haowen Deng, Tolga Birdal, and Slobodan Ilic.
\newblock Ppfnet: Global context aware local features for robust 3d point matching.
\newblock In \emph{Proceedings of the IEEE conference on computer vision and pattern recognition}, pages 195--205, 2018.

\bibitem[Deng et~al.(2022)Deng, Liu, Zhu, and Ramanan]{deng2022depth}
Kangle Deng, Andrew Liu, Jun-Yan Zhu, and Deva Ramanan.
\newblock Depth-supervised nerf: Fewer views and faster training for free.
\newblock In \emph{Proceedings of the IEEE/CVF Conference on Computer Vision and Pattern Recognition}, pages 12882--12891, 2022.

\bibitem[Dou et~al.(2016)Dou, Khamis, Degtyarev, Davidson, Fanello, Kowdle, Escolano, Rhemann, Kim, Taylor, et~al.]{dou2016fusion4d}
Mingsong Dou, Sameh Khamis, Yury Degtyarev, Philip Davidson, Sean~Ryan Fanello, Adarsh Kowdle, Sergio~Orts Escolano, Christoph Rhemann, David Kim, Jonathan Taylor, et~al.
\newblock Fusion4d: Real-time performance capture of challenging scenes.
\newblock \emph{ACM Transactions on Graphics (ToG)}, 35\penalty0 (4):\penalty0 1--13, 2016.

\bibitem[Drost et~al.(2010)Drost, Ulrich, Navab, and Ilic]{drost2010model}
Bertram Drost, Markus Ulrich, Nassir Navab, and Slobodan Ilic.
\newblock Model globally, match locally: Efficient and robust 3d object recognition.
\newblock In \emph{2010 IEEE computer society conference on computer vision and pattern recognition}, pages 998--1005. Ieee, 2010.

\bibitem[G{\"u}ler et~al.(2018)G{\"u}ler, Neverova, and Kokkinos]{guler2018densepose}
R{\i}za~Alp G{\"u}ler, Natalia Neverova, and Iasonas Kokkinos.
\newblock Densepose: Dense human pose estimation in the wild.
\newblock In \emph{Proceedings of the IEEE conference on computer vision and pattern recognition}, pages 7297--7306, 2018.

\bibitem[Guo et~al.(2020)Guo, Wang, Hu, Liu, Liu, and Bennamoun]{guo2020deep}
Yulan Guo, Hanyun Wang, Qingyong Hu, Hao Liu, Li Liu, and Mohammed Bennamoun.
\newblock Deep learning for 3d point clouds: A survey.
\newblock \emph{IEEE transactions on pattern analysis and machine intelligence}, 43\penalty0 (12):\penalty0 4338--4364, 2020.

\bibitem[Habermann et~al.(2020)Habermann, Xu, Zollhofer, Pons-Moll, and Theobalt]{habermann2020deepcap}
Marc Habermann, Weipeng Xu, Michael Zollhofer, Gerard Pons-Moll, and Christian Theobalt.
\newblock Deepcap: Monocular human performance capture using weak supervision.
\newblock In \emph{Proceedings of the IEEE/CVF Conference on Computer Vision and Pattern Recognition}, pages 5052--5063, 2020.

\bibitem[Insafutdinov et~al.(2016)Insafutdinov, Pishchulin, Andres, Andriluka, and Schiele]{insafutdinov2016deepercut}
Eldar Insafutdinov, Leonid Pishchulin, Bjoern Andres, Mykhaylo Andriluka, and Bernt Schiele.
\newblock Deepercut: A deeper, stronger, and faster multi-person pose estimation model.
\newblock In \emph{Computer Vision--ECCV 2016: 14th European Conference, Amsterdam, The Netherlands, October 11-14, 2016, Proceedings, Part VI 14}, pages 34--50. Springer, 2016.

\bibitem[Ionescu et~al.(2013)Ionescu, Papava, Olaru, and Sminchisescu]{ionescu2013human3p6}
Catalin Ionescu, Dragos Papava, Vlad Olaru, and Cristian Sminchisescu.
\newblock Human3. 6m: Large scale datasets and predictive methods for 3d human sensing in natural environments.
\newblock \emph{IEEE transactions on pattern analysis and machine intelligence}, 36\penalty0 (7):\penalty0 1325--1339, 2013.

\bibitem[I{\c{s}}{\i}k et~al.(2023)I{\c{s}}{\i}k, R{\"u}nz, Georgopoulos, Khakhulin, Starck, Agapito, and Nie{\ss}ner]{icsik2023humanrf}
Mustafa I{\c{s}}{\i}k, Martin R{\"u}nz, Markos Georgopoulos, Taras Khakhulin, Jonathan Starck, Lourdes Agapito, and Matthias Nie{\ss}ner.
\newblock Humanrf: High-fidelity neural radiance fields for humans in motion.
\newblock \emph{arXiv preprint arXiv:2305.06356}, 2023.

\bibitem[Jiang et~al.(2022{\natexlab{a}})Jiang, Hong, Bao, and Zhang]{jiang2022selfrecon}
Boyi Jiang, Yang Hong, Hujun Bao, and Juyong Zhang.
\newblock Selfrecon: Self reconstruction your digital avatar from monocular video.
\newblock In \emph{Proceedings of the IEEE/CVF Conference on Computer Vision and Pattern Recognition}, pages 5605--5615, 2022{\natexlab{a}}.

\bibitem[Jiang et~al.(2022{\natexlab{b}})Jiang, Yi, Samei, Tuzel, and Ranjan]{jiang2022neuman}
Wei Jiang, Kwang~Moo Yi, Golnoosh Samei, Oncel Tuzel, and Anurag Ranjan.
\newblock Neuman: Neural human radiance field from a single video.
\newblock In \emph{European Conference on Computer Vision}, pages 402--418. Springer, 2022{\natexlab{b}}.

\bibitem[Jung et~al.(2022)Jung, Wu, Ruhkamp, Schieber, Wang, Rizzoli, Zhao, Meier, Roth, Navab, et~al.]{jung2022housecat6d}
HyunJun Jung, Shun-Cheng Wu, Patrick Ruhkamp, Hannah Schieber, Pengyuan Wang, Giulia Rizzoli, Hongcheng Zhao, Sven~Damian Meier, Daniel Roth, Nassir Navab, et~al.
\newblock Housecat6d--a large-scale multi-modal category level 6d object pose dataset with household objects in realistic scenarios.
\newblock \emph{arXiv preprint arXiv:2212.10428}, 2022.

\bibitem[Jung et~al.(2023)Jung, Ruhkamp, Zhai, Brasch, Li, Verdie, Song, Zhou, Armagan, Ilic, et~al.]{jung2023importance}
HyunJun Jung, Patrick Ruhkamp, Guangyao Zhai, Nikolas Brasch, Yitong Li, Yannick Verdie, Jifei Song, Yiren Zhou, Anil Armagan, Slobodan Ilic, et~al.
\newblock On the importance of accurate geometry data for dense 3d vision tasks.
\newblock In \emph{Proceedings of the IEEE/CVF Conference on Computer Vision and Pattern Recognition}, pages 780--791, 2023.

\bibitem[Kanade et~al.(1997)Kanade, Rander, and Narayanan]{kanade1997virtualized}
Takeo Kanade, Peter Rander, and PJ Narayanan.
\newblock Virtualized reality: Constructing virtual worlds from real scenes.
\newblock \emph{IEEE multimedia}, 4\penalty0 (1):\penalty0 34--47, 1997.

\bibitem[Karaoglu et~al.(2023)Karaoglu, Schieber, Schischka, G{\"o}rg{\"u}l{\"u}, Gr{\"o}tzner, Ladikos, Roth, Navab, and Busam]{karaoglu2023dynamon}
Mert~Asim Karaoglu, Hannah Schieber, Nicolas Schischka, Melih G{\"o}rg{\"u}l{\"u}, Florian Gr{\"o}tzner, Alexander Ladikos, Daniel Roth, Nassir Navab, and Benjamin Busam.
\newblock Dynamon: Motion-aware fast and robust camera localization for dynamic nerf.
\newblock \emph{arXiv preprint arXiv:2309.08927}, 2023.

\bibitem[Kerbl et~al.(2023)Kerbl, Kopanas, Leimk{\"u}hler, and Drettakis]{kerbl20233d}
Bernhard Kerbl, Georgios Kopanas, Thomas Leimk{\"u}hler, and George Drettakis.
\newblock 3d gaussian splatting for real-time radiance field rendering.
\newblock \emph{ACM Transactions on Graphics (ToG)}, 42\penalty0 (4):\penalty0 1--14, 2023.

\bibitem[Krizhevsky et~al.(2012)Krizhevsky, Sutskever, and Hinton]{Krizhevsky2012ImageNetCW}
Alex Krizhevsky, Ilya Sutskever, and Geoffrey~E. Hinton.
\newblock Imagenet classification with deep convolutional neural networks.
\newblock \emph{Communications of the ACM}, 60:\penalty0 84 -- 90, 2012.

\bibitem[Levoy and Hanrahan(1996)]{levoy1996lightfield}
Marc Levoy and Pat Hanrahan.
\newblock Light field rendering.
\newblock In \emph{Proceedings of the 23rd Annual Conference on Computer Graphics and Interactive Techniques}, page 31–42, New York, NY, USA, 1996. Association for Computing Machinery.

\bibitem[Li et~al.(2022{\natexlab{a}})Li, Tanke, Vo, Zollh{\"o}fer, Gall, Kanazawa, and Lassner]{li2022tava}
Ruilong Li, Julian Tanke, Minh Vo, Michael Zollh{\"o}fer, J{\"u}rgen Gall, Angjoo Kanazawa, and Christoph Lassner.
\newblock Tava: Template-free animatable volumetric actors.
\newblock In \emph{European Conference on Computer Vision}, pages 419--436. Springer, 2022{\natexlab{a}}.

\bibitem[Li et~al.(2022{\natexlab{b}})Li, Slavcheva, Zollhoefer, Green, Lassner, Kim, Schmidt, Lovegrove, Goesele, Newcombe, et~al.]{li2022neural}
Tianye Li, Mira Slavcheva, Michael Zollhoefer, Simon Green, Christoph Lassner, Changil Kim, Tanner Schmidt, Steven Lovegrove, Michael Goesele, Richard Newcombe, et~al.
\newblock Neural 3d video synthesis from multi-view video.
\newblock In \emph{Proceedings of the IEEE/CVF Conference on Computer Vision and Pattern Recognition}, pages 5521--5531, 2022{\natexlab{b}}.

\bibitem[Li et~al.(2023)Li, Zheng, Liu, Zhou, and Liu]{li2023posevocab}
Zhe Li, Zerong Zheng, Yuxiao Liu, Boyao Zhou, and Yebin Liu.
\newblock Posevocab: Learning joint-structured pose embeddings for human avatar modeling.
\newblock In \emph{ACM SIGGRAPH Conference Proceedings}, 2023.

\bibitem[Lin et~al.(2021)Lin, Ma, Torralba, and Lucey]{lin2021barf}
Chen-Hsuan Lin, Wei-Chiu Ma, Antonio Torralba, and Simon Lucey.
\newblock Barf: Bundle-adjusting neural radiance fields.
\newblock In \emph{Proceedings of the IEEE/CVF International Conference on Computer Vision}, pages 5741--5751, 2021.

\bibitem[Litany et~al.(2017)Litany, Remez, Rodola, Bronstein, and Bronstein]{litany2017deep}
Or Litany, Tal Remez, Emanuele Rodola, Alex Bronstein, and Michael Bronstein.
\newblock Deep functional maps: Structured prediction for dense shape correspondence.
\newblock In \emph{Proceedings of the IEEE international conference on computer vision}, pages 5659--5667, 2017.

\bibitem[Liu et~al.(2021)Liu, Habermann, Rudnev, Sarkar, Gu, and Theobalt]{liu2021neural}
Lingjie Liu, Marc Habermann, Viktor Rudnev, Kripasindhu Sarkar, Jiatao Gu, and Christian Theobalt.
\newblock Neural actor: Neural free-view synthesis of human actors with pose control.
\newblock \emph{ACM transactions on graphics (TOG)}, 40\penalty0 (6):\penalty0 1--16, 2021.

\bibitem[Loper et~al.(2015)Loper, Mahmood, Romero, Pons-Moll, and Black]{loper2015smpl}
Matthew Loper, Naureen Mahmood, Javier Romero, Gerard Pons-Moll, and Michael~J Black.
\newblock Smpl: A skinned multi-person linear model.
\newblock \emph{ACM transactions on graphics (TOG)}, 34\penalty0 (6):\penalty0 1--16, 2015.

\bibitem[Luiten et~al.(2024)Luiten, Kopanas, Leibe, and Ramanan]{luiten2023dynamic}
Jonathon Luiten, Georgios Kopanas, Bastian Leibe, and Deva Ramanan.
\newblock Dynamic 3d gaussians: Tracking by persistent dynamic view synthesis.
\newblock In \emph{3DV}, 2024.

\bibitem[Manhardt et~al.(2019)Manhardt, Arroyo, Rupprecht, Busam, Birdal, Navab, and Tombari]{manhardt2019explaining}
Fabian Manhardt, Diego~Martin Arroyo, Christian Rupprecht, Benjamin Busam, Tolga Birdal, Nassir Navab, and Federico Tombari.
\newblock Explaining the ambiguity of object detection and 6d pose from visual data.
\newblock In \emph{Proceedings of the IEEE/CVF International Conference on Computer Vision}, pages 6841--6850, 2019.

\bibitem[Mescheder et~al.(2019)Mescheder, Oechsle, Niemeyer, Nowozin, and Geiger]{Mescheder2019occupancy}
Lars Mescheder, Michael Oechsle, Michael Niemeyer, Sebastian Nowozin, and Andreas Geiger.
\newblock Occupancy networks: Learning 3d reconstruction in function space.
\newblock In \emph{Proceedings of the IEEE/CVF Conference on Computer Vision and Pattern Recognition (CVPR)}, 2019.

\bibitem[Mildenhall et~al.(2021)Mildenhall, Srinivasan, Tancik, Barron, Ramamoorthi, and Ng]{mildenhall2021nerf}
Ben Mildenhall, Pratul~P Srinivasan, Matthew Tancik, Jonathan~T Barron, Ravi Ramamoorthi, and Ren Ng.
\newblock Nerf: Representing scenes as neural radiance fields for view synthesis.
\newblock \emph{Communications of the ACM}, 65\penalty0 (1):\penalty0 99--106, 2021.

\bibitem[M{\"u}ller et~al.(2022)M{\"u}ller, Evans, Schied, and Keller]{muller2022instant}
Thomas M{\"u}ller, Alex Evans, Christoph Schied, and Alexander Keller.
\newblock Instant neural graphics primitives with a multiresolution hash encoding.
\newblock \emph{ACM Transactions on Graphics (ToG)}, 41\penalty0 (4):\penalty0 1--15, 2022.

\bibitem[Noguchi et~al.(2021)Noguchi, Sun, Lin, and Harada]{noguchi2021neural}
Atsuhiro Noguchi, Xiao Sun, Stephen Lin, and Tatsuya Harada.
\newblock Neural articulated radiance field.
\newblock In \emph{Proceedings of the IEEE/CVF International Conference on Computer Vision}, pages 5762--5772, 2021.

\bibitem[Orts-Escolano et~al.(2016)Orts-Escolano, Rhemann, Fanello, Chang, Kowdle, Degtyarev, Kim, Davidson, Khamis, Dou, et~al.]{orts2016holoportation}
Sergio Orts-Escolano, Christoph Rhemann, Sean Fanello, Wayne Chang, Adarsh Kowdle, Yury Degtyarev, David Kim, Philip~L Davidson, Sameh Khamis, Mingsong Dou, et~al.
\newblock Holoportation: Virtual 3d teleportation in real-time.
\newblock In \emph{Proceedings of the 29th annual symposium on user interface software and technology}, pages 741--754, 2016.

\bibitem[Park et~al.(2019)Park, Florence, Straub, Newcombe, and Lovegrove]{Park2019deepsdf}
Jeong~Joon Park, Peter Florence, Julian Straub, Richard Newcombe, and Steven Lovegrove.
\newblock Deepsdf: Learning continuous signed distance functions for shape representation.
\newblock In \emph{Proceedings of the IEEE/CVF Conference on Computer Vision and Pattern Recognition (CVPR)}, 2019.

\bibitem[Park et~al.(2021{\natexlab{a}})Park, Sinha, Barron, Bouaziz, Goldman, Seitz, and Martin-Brualla]{park2021nerfies}
Keunhong Park, Utkarsh Sinha, Jonathan~T Barron, Sofien Bouaziz, Dan~B Goldman, Steven~M Seitz, and Ricardo Martin-Brualla.
\newblock Nerfies: Deformable neural radiance fields.
\newblock In \emph{Proceedings of the IEEE/CVF International Conference on Computer Vision}, pages 5865--5874, 2021{\natexlab{a}}.

\bibitem[Park et~al.(2021{\natexlab{b}})Park, Sinha, Hedman, Barron, Bouaziz, Goldman, Martin-Brualla, and Seitz]{park2021hypernerf}
Keunhong Park, Utkarsh Sinha, Peter Hedman, Jonathan~T Barron, Sofien Bouaziz, Dan~B Goldman, Ricardo Martin-Brualla, and Steven~M Seitz.
\newblock Hypernerf: A higher-dimensional representation for topologically varying neural radiance fields.
\newblock \emph{arXiv preprint arXiv:2106.13228}, 2021{\natexlab{b}}.

\bibitem[Peng et~al.(2021{\natexlab{a}})Peng, Dong, Wang, Zhang, Shuai, Zhou, and Bao]{peng2021animatable}
Sida Peng, Junting Dong, Qianqian Wang, Shangzhan Zhang, Qing Shuai, Xiaowei Zhou, and Hujun Bao.
\newblock Animatable neural radiance fields for modeling dynamic human bodies.
\newblock In \emph{Proceedings of the IEEE/CVF International Conference on Computer Vision}, pages 14314--14323, 2021{\natexlab{a}}.

\bibitem[Peng et~al.(2021{\natexlab{b}})Peng, Zhang, Xu, Wang, Shuai, Bao, and Zhou]{peng2021neural}
Sida Peng, Yuanqing Zhang, Yinghao Xu, Qianqian Wang, Qing Shuai, Hujun Bao, and Xiaowei Zhou.
\newblock Neural body: Implicit neural representations with structured latent codes for novel view synthesis of dynamic humans.
\newblock In \emph{CVPR}, 2021{\natexlab{b}}.

\bibitem[Pfister et~al.(2000)Pfister, Zwicker, van Baar, and Gross]{sufel}
Hanspeter Pfister, Matthias Zwicker, Jeroen van Baar, and Markus Gross.
\newblock Surfels: Surface elements as rendering primitives.
\newblock In \emph{Proceedings of the 27th Annual Conference on Computer Graphics and Interactive Techniques}, page 335–342, USA, 2000. ACM Press/Addison-Wesley Publishing Co.

\bibitem[Pumarola et~al.(2021)Pumarola, Corona, Pons-Moll, and Moreno-Noguer]{pumarola2021d}
Albert Pumarola, Enric Corona, Gerard Pons-Moll, and Francesc Moreno-Noguer.
\newblock D-nerf: Neural radiance fields for dynamic scenes.
\newblock In \emph{Proceedings of the IEEE/CVF Conference on Computer Vision and Pattern Recognition}, pages 10318--10327, 2021.

\bibitem[Qin et~al.(2022)Qin, Yu, Wang, Guo, Peng, and Xu]{qin2022geometric}
Zheng Qin, Hao Yu, Changjian Wang, Yulan Guo, Yuxing Peng, and Kai Xu.
\newblock Geometric transformer for fast and robust point cloud registration.
\newblock In \emph{Proceedings of the IEEE/CVF conference on computer vision and pattern recognition}, pages 11143--11152, 2022.

\bibitem[Roessle et~al.(2022)Roessle, Barron, Mildenhall, Srinivasan, and Nie{\ss}ner]{roessle2022dense}
Barbara Roessle, Jonathan~T Barron, Ben Mildenhall, Pratul~P Srinivasan, and Matthias Nie{\ss}ner.
\newblock Dense depth priors for neural radiance fields from sparse input views.
\newblock In \emph{Proceedings of the IEEE/CVF Conference on Computer Vision and Pattern Recognition}, pages 12892--12901, 2022.

\bibitem[Rusu and Cousins(2011)]{rusu20113d}
Radu~Bogdan Rusu and Steve Cousins.
\newblock 3d is here: Point cloud library (pcl).
\newblock In \emph{2011 IEEE international conference on robotics and automation}, pages 1--4. IEEE, 2011.

\bibitem[Saito et~al.(2021)Saito, Yang, Ma, and Black]{saito2021scanimate}
Shunsuke Saito, Jinlong Yang, Qianli Ma, and Michael~J Black.
\newblock Scanimate: Weakly supervised learning of skinned clothed avatar networks.
\newblock In \emph{Proceedings of the IEEE/CVF Conference on Computer Vision and Pattern Recognition}, pages 2886--2897, 2021.

\bibitem[Saleh et~al.(2022)Saleh, Wu, Cosmo, Navab, Busam, and Tombari]{saleh2022bending}
Mahdi Saleh, Shun-Cheng Wu, Luca Cosmo, Nassir Navab, Benjamin Busam, and Federico Tombari.
\newblock Bending graphs: Hierarchical shape matching using gated optimal transport.
\newblock In \emph{Proceedings of the IEEE/CVF Conference on Computer Vision and Pattern Recognition}, pages 11757--11767, 2022.

\bibitem[Salti et~al.(2014)Salti, Tombari, and Di~Stefano]{salti2014shot}
Samuele Salti, Federico Tombari, and Luigi Di~Stefano.
\newblock Shot: Unique signatures of histograms for surface and texture description.
\newblock \emph{Computer Vision and Image Understanding}, 125:\penalty0 251--264, 2014.

\bibitem[Sch\"{o}nberger and Frahm(2016)]{schoenberger2016sfm}
Johannes~Lutz Sch\"{o}nberger and Jan-Michael Frahm.
\newblock Structure-from-motion revisited.
\newblock In \emph{Conference on Computer Vision and Pattern Recognition (CVPR)}, 2016.

\bibitem[Sch\"{o}nberger et~al.(2016)Sch\"{o}nberger, Zheng, Pollefeys, and Frahm]{schoenberger2016mvs}
Johannes~Lutz Sch\"{o}nberger, Enliang Zheng, Marc Pollefeys, and Jan-Michael Frahm.
\newblock Pixelwise view selection for unstructured multi-view stereo.
\newblock In \emph{European Conference on Computer Vision (ECCV)}, 2016.

\bibitem[Seitz and Dyer(1999)]{seitz1999photorealistic}
Steven~M Seitz and Charles~R Dyer.
\newblock Photorealistic scene reconstruction by voxel coloring.
\newblock \emph{International Journal of Computer Vision}, 35:\penalty0 151--173, 1999.

\bibitem[Song et~al.(2023)Song, Chen, Li, Chen, Chen, Yuan, Xu, and Geiger]{Song2023nerfplayer}
Liangchen Song, Anpei Chen, Zhong Li, Zhang Chen, Lele Chen, Junsong Yuan, Yi Xu, and Andreas Geiger.
\newblock Nerfplayer: A streamable dynamic scene representation with decomposed neural radiance fields.
\newblock \emph{IEEE Transactions on Visualization and Computer Graphics}, 29\penalty0 (5):\penalty0 2732--2742, 2023.

\bibitem[Starck and Hilton(2007)]{starck2007surface}
Jonathan Starck and Adrian Hilton.
\newblock Surface capture for performance-based animation.
\newblock \emph{IEEE computer graphics and applications}, 27\penalty0 (3):\penalty0 21--31, 2007.

\bibitem[Su et~al.(2021)Su, Yu, Zollh{\"o}fer, and Rhodin]{su2021nerf}
Shih-Yang Su, Frank Yu, Michael Zollh{\"o}fer, and Helge Rhodin.
\newblock A-nerf: Articulated neural radiance fields for learning human shape, appearance, and pose.
\newblock \emph{Advances in Neural Information Processing Systems}, 34:\penalty0 12278--12291, 2021.

\bibitem[Svitov et~al.(2023)Svitov, Gudkov, Bashirov, and Lempitsky]{svitov2023dinar}
David Svitov, Dmitrii Gudkov, Renat Bashirov, and Victor Lempitsky.
\newblock Dinar: Diffusion inpainting of neural textures for one-shot human avatars.
\newblock In \emph{Proceedings of the IEEE/CVF International Conference on Computer Vision}, pages 7062--7072, 2023.

\bibitem[Szeliski and Golland(1998)]{szeliski1998stereo}
Richard Szeliski and Polina Golland.
\newblock Stereo matching with transparency and matting.
\newblock In \emph{Sixth International Conference on Computer Vision (IEEE Cat. No. 98CH36271)}, pages 517--524. IEEE, 1998.

\bibitem[Szeliski et~al.(1996)Szeliski, Gortler, Grzeszczuk, and Cohen]{szeliski1996lumigraph}
Richard Szeliski, Steven Gortler, Radek Grzeszczuk, and Michael~F Cohen.
\newblock The lumigraph.
\newblock In \emph{Proceedings of the 23rd annual conference on computer graphics and interactive techniques (SIGGRAPH 1996)}, pages 43--54, 1996.

\bibitem[Wang et~al.(2022{\natexlab{a}})Wang, Jung, Li, Shen, Srikanth, Garattoni, Meier, Navab, and Busam]{wang2022phocal}
Pengyuan Wang, HyunJun Jung, Yitong Li, Siyuan Shen, Rahul~Parthasarathy Srikanth, Lorenzo Garattoni, Sven Meier, Nassir Navab, and Benjamin Busam.
\newblock Phocal: A multi-modal dataset for category-level object pose estimation with photometrically challenging objects.
\newblock In \emph{Proceedings of the IEEE/CVF Conference on Computer Vision and Pattern Recognition}, pages 21222--21231, 2022{\natexlab{a}}.

\bibitem[Wang et~al.(2022{\natexlab{b}})Wang, Schwarz, Geiger, and Tang]{wang2022arah}
Shaofei Wang, Katja Schwarz, Andreas Geiger, and Siyu Tang.
\newblock Arah: Animatable volume rendering of articulated human sdfs.
\newblock In \emph{European conference on computer vision}, pages 1--19. Springer, 2022{\natexlab{b}}.

\bibitem[Wang et~al.(2021{\natexlab{a}})Wang, Li, Zhang, and Shan]{wang2021towards}
Xintao Wang, Yu Li, Honglun Zhang, and Ying Shan.
\newblock Towards real-world blind face restoration with generative facial prior.
\newblock In \emph{Proceedings of the IEEE/CVF conference on computer vision and pattern recognition}, pages 9168--9178, 2021{\natexlab{a}}.

\bibitem[Wang et~al.(2021{\natexlab{b}})Wang, Wu, Xie, Chen, and Prisacariu]{wang2021nerf}
Zirui Wang, Shangzhe Wu, Weidi Xie, Min Chen, and Victor~Adrian Prisacariu.
\newblock Nerf--: Neural radiance fields without known camera parameters.
\newblock \emph{arXiv preprint arXiv:2102.07064}, 2021{\natexlab{b}}.

\bibitem[Weng et~al.(2022)Weng, Curless, Srinivasan, Barron, and Kemelmacher-Shlizerman]{weng2022humannerf}
Chung-Yi Weng, Brian Curless, Pratul~P Srinivasan, Jonathan~T Barron, and Ira Kemelmacher-Shlizerman.
\newblock Humannerf: Free-viewpoint rendering of moving people from monocular video.
\newblock In \emph{Proceedings of the IEEE/CVF conference on computer vision and pattern Recognition}, pages 16210--16220, 2022.

\bibitem[Wu et~al.(2023)Wu, Yi, Fang, Xie, Zhang, Wei, Liu, Tian, and Wang]{wu20234d}
Guanjun Wu, Taoran Yi, Jiemin Fang, Lingxi Xie, Xiaopeng Zhang, Wei Wei, Wenyu Liu, Qi Tian, and Xinggang Wang.
\newblock 4d gaussian splatting for real-time dynamic scene rendering.
\newblock \emph{arXiv preprint arXiv:2310.08528}, 2023.

\bibitem[Xu et~al.(2022)Xu, Fujita, and Matsumoto]{xu2022surface}
Tianhan Xu, Yasuhiro Fujita, and Eiichi Matsumoto.
\newblock Surface-aligned neural radiance fields for controllable 3d human synthesis.
\newblock In \emph{Proceedings of the IEEE/CVF Conference on Computer Vision and Pattern Recognition}, pages 15883--15892, 2022.

\bibitem[Xu et~al.(2018)Xu, Chatterjee, Zollh{\"o}fer, Rhodin, Mehta, Seidel, and Theobalt]{xu2018monoperfcap}
Weipeng Xu, Avishek Chatterjee, Michael Zollh{\"o}fer, Helge Rhodin, Dushyant Mehta, Hans-Peter Seidel, and Christian Theobalt.
\newblock Monoperfcap: Human performance capture from monocular video.
\newblock \emph{ACM Transactions on Graphics (ToG)}, 37\penalty0 (2):\penalty0 1--15, 2018.

\bibitem[Xu et~al.(2023)Xu, Peng, Lin, He, Sun, Shen, Bao, and Zhou]{xu20234k4d}
Zhen Xu, Sida Peng, Haotong Lin, Guangzhao He, Jiaming Sun, Yujun Shen, Hujun Bao, and Xiaowei Zhou.
\newblock 4k4d: Real-time 4d view synthesis at 4k resolution, 2023.

\bibitem[Yang et~al.(2023)Yang, Gao, Zhou, Jiao, Zhang, and Jin]{yang2023deformable3dgs}
Ziyi Yang, Xinyu Gao, Wen Zhou, Shaohui Jiao, Yuqing Zhang, and Xiaogang Jin.
\newblock Deformable 3d gaussians for high-fidelity monocular dynamic scene reconstruction.
\newblock \emph{arXiv preprint arXiv:2309.13101}, 2023.

\bibitem[Yu et~al.(2021)Yu, Li, Saleh, Busam, and Ilic]{yu2021cofinet}
Hao Yu, Fu Li, Mahdi Saleh, Benjamin Busam, and Slobodan Ilic.
\newblock Cofinet: Reliable coarse-to-fine correspondences for robust pointcloud registration.
\newblock \emph{Advances in Neural Information Processing Systems}, 34:\penalty0 23872--23884, 2021.

\bibitem[Yu et~al.(2023{\natexlab{a}})Yu, Qin, Hou, Saleh, Li, Busam, and Ilic]{yu2023rotation}
Hao Yu, Zheng Qin, Ji Hou, Mahdi Saleh, Dongsheng Li, Benjamin Busam, and Slobodan Ilic.
\newblock Rotation-invariant transformer for point cloud matching.
\newblock In \emph{Proceedings of the IEEE/CVF Conference on Computer Vision and Pattern Recognition}, pages 5384--5393, 2023{\natexlab{a}}.

\bibitem[Yu et~al.(2023{\natexlab{b}})Yu, Cheng, Liu, Wu, and Lin]{Yu_2023_CVPR}
Zhengming Yu, Wei Cheng, Xian Liu, Wayne Wu, and Kwan-Yee Lin.
\newblock Monohuman: Animatable human neural field from monocular video.
\newblock In \emph{Proceedings of the IEEE/CVF Conference on Computer Vision and Pattern Recognition (CVPR)}, pages 16943--16953, 2023{\natexlab{b}}.

\bibitem[Zheng et~al.(2022)Zheng, Huang, Yu, Zhang, Guo, and Liu]{zheng2022structured}
Zerong Zheng, Han Huang, Tao Yu, Hongwen Zhang, Yandong Guo, and Yebin Liu.
\newblock Structured local radiance fields for human avatar modeling.
\newblock In \emph{Proceedings of the IEEE/CVF Conference on Computer Vision and Pattern Recognition (CVPR)}, 2022.

\bibitem[Zielonka et~al.(2023)Zielonka, Bagautdinov, Saito, Zollhöfer, Thies, and Romero]{zielonka2023drivable}
Wojciech Zielonka, Timur Bagautdinov, Shunsuke Saito, Michael Zollhöfer, Justus Thies, and Javier Romero.
\newblock Drivable 3d gaussian avatars, 2023.

\end{thebibliography}
}

\clearpage
\setcounter{page}{1}
\maketitlesupplementary

\section{Deformation Refinement Module (DRM)}
\begin{figure}[!hb]
 \centering
    \includegraphics[width=1.0\linewidth]{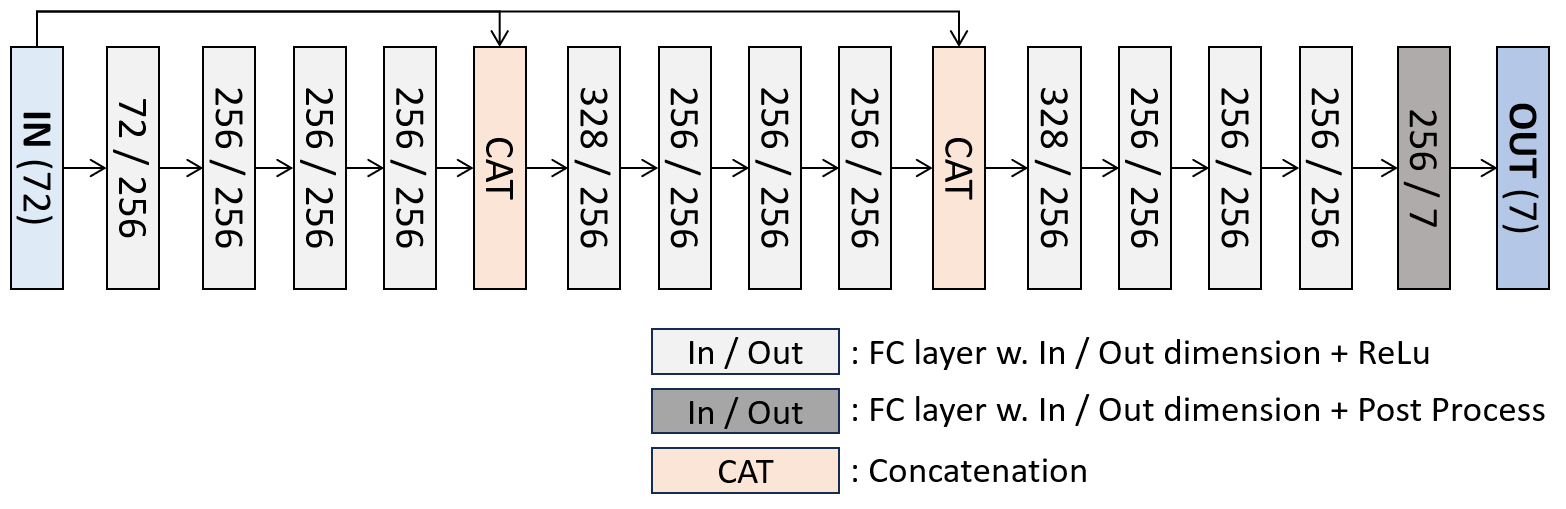}
    \caption{\textbf{DRM definition} The DRM is based on 13 FC layers. Each of the 12 layers is followed by a ReLu activation and the last layer is followed by a postprocessing that converts the 7 channel vector into translation and rotation. 
    The 5th and 9th layer are skip connection layers that concatenate the previous layer's output feature and input vector.}
    \label{fig:drm_layer}
\end{figure}

The DRM is a fully connected (FC) layer based network that takes care of the motion dependent cloth deformation in our pipeline. The DRM is composed of 13 FC layers. The first 12 FC layers are followed by a ReLu activation. After the 5th and 9th layer there are skip connections. The last layer is a FC layer followed by a postprocessing layer to convert the output vector into rotation and translation. A detailed composition is shown in Fig.~\ref{fig:drm_layer}.

The last layer outputs a 7 channel vector that is parameterized as axis angle rotation (4 channels) and translation (3 channels). We use the first 3 channels as translation and the later channels for rotation. For the rotation, the first channel is used as the angle, and the later 3 channels are normalized to form the rotation axis. For the ZJU mocap dataset~\cite{peng2021neural}, we apply a sigmoid activiation to limit the translation to \(\pm 10cm\) and rotation to \(\pm 30^{\circ}\). For the THUman 4.0 dataset~\cite{zheng2022structured}, we use an unbounded output that is scaled by 0.01 for the translation \((1 \to 1cm)\) and \(\pi^{-1}\) for the rotation \((1 \to 18.25^{\circ})\). We find the results for the THUman dataset~\cite{zheng2022structured} are better with an unbounded prediction as the human performers wears hoodies that involve more deformation than the ZJU~\cite{peng2021neural} dataset.

\section{Camera Selection for Training and Evaluation}
We use 7 camera views for training for both the ZJU mocap dataset~\cite{peng2021neural} and the THUman 4.0 dataset~\cite{zheng2022structured}. Tab.~\ref{tab:camera_index} shows the exact cameras that are used for training and testing. For the ZJU mocap dataset~\cite{peng2021neural}, slightly different training and testing cameras are used depending on the scene due to issues with camera calibration (see Fig.~\ref{fig:camera_issue} (a)). While for the THUman 4.0 dataset~\cite{zheng2022structured}, we fix the cameras among all scenes for training and testing. For both dataset, we use mixed cameras for testing. For the ZJU mocap dataset~\cite{peng2021neural}, 3 seen views and 4 unseen views are used for testing, while 3 seen and 3 unseen views are used for the THUman 4.0 dataset~\cite{zheng2022structured}.

\begin{table}[h!]
\caption{\textbf{Cameras Used for Training and Testing.} We use different cameras in the ZJU mocap dataset~\cite{peng2021neural} among the scenes as the calibration quality differs depending on the scene. In comparison, the same camera combinations are used in all scenes in THUman 4.0 dataset~\cite{zheng2022structured}. For testing we use a combination of seen and unseen cameras for both datasets.}
\label{tab:camera_index}
\resizebox{\columnwidth}{!}{
\begin{tabular}{c|c|c|c}
\hline
Dataset                     & Scene & Train Camera      & Test Camera        \\ \hline
\multirow{4}{*}{ZJU Mocap~\cite{peng2021neural}}  & 313   & 1,3,7,10,13,16,19 & 1,5,10,13,16,18,21 \\
                            & 315   & 1,3,7,10,13,16,19 & 1,5,10,13,16,18,21 \\
                            & 377   & 2,3,7,10,14,15,16 & 2,4,7,10,14,17,22  \\
                            & 394   & 2,5,7,10,15,16,22 & 3,4,7,10,15,16,23  \\ \hline
\multirow{3}{*}{THUman 4.0~\cite{zheng2022structured}} & 00    & 1,5,9,13,17,21,23 & 3,7,11,17,21,23    \\
                            & 01    & 1,5,9,13,17,21,23 & 3,7,11,17,21,23    \\
                            & 02    & 1,5,9,13,17,21,23 & 3,7,11,17,21,23    \\ \hline
\end{tabular}
}
\end{table}


\begin{table*}[]
\caption{\textbf{Detailed time analysis.} We break down each step of our pipeline during inference to give detailed information regarding speed and also insight for future optimization. In general, generating an human avatar from the THUman dataset~\cite{zheng2022structured} takes more time than from the Mocap dataset~\cite{peng2021neural}. We find that, short sleeves and short pants require less gaussians than long sleeves, hoodies and long pants. For this reason, the rendering speed on scene 313-377 is slightly faster than scene 394, and the Mocap dataset~\cite{peng2021neural} is faster than the THUman dataset~\cite{zheng2022structured} on average.}
\label{tab:time_analysis}
\resizebox{\textwidth}{!}{
\begin{tabular}{c|cccccc|c|c}
\hline
Scene        & \begin{tabular}[c]{@{}c@{}}Per Vertrex\\ Deformation\\ Calculation\end{tabular} & Posing (I)      & \begin{tabular}[c]{@{}c@{}}Deformation\\ Refinement\\ Calculation\end{tabular} & Posing (II)     & Rendering       & Img Save       & \begin{tabular}[c]{@{}c@{}}Total Time\\ (w. Save / w.o. Save)\end{tabular} & \begin{tabular}[c]{@{}c@{}}FPS\\ (w. Save / w.o. Save)\end{tabular} \\ \hline
313          & 0.0414                                                                          & 0.0069          & 0.0479                                                                         & 0.0026          & 0.0068          & 0.041          & 0.147 / 0.106                                                              & 6.80 / 9.43                                                         \\
315          & 0.0437                                                                          & 0.0074          & 0.0536                                                                         & 0.0028          & 0.0072          & 0.042          & 0.158 / 0.116                                                              & 6.33 / 8.62                                                         \\
377          & 0.0434                                                                          & 0.0080          & 0.0434                                                                         & 0.0030          & 0.0074          & 0.043          & 0.148 / 0.105                                                              & 6.75 / 9.53                                                         \\
394          & 0.0410                                                                          & 0.0093          & 0.0568                                                                         & 0.0034          & 0.0084          & 0.042          & 0.161 / 0.119                                                              & 6.21 / 8.40                                                         \\
\textbf{AVG} & \textbf{0.0424}                                                                 & \textbf{0.0079} & \textbf{0.0504}                                                                & \textbf{0.0030} & \textbf{0.0075} & \textbf{0.042} & \textbf{0.153 / 0.111}                                                     & \textbf{6.54 / 9.01}                                                \\ \hline
00    & 0.0652                                                                          & 0.0041          & 0.0369                                                                         & 0.0017          & 0.0091          & 0.049          & 0.166 / 0.117                                                              & 6.02 / 8.55                                                         \\
01    & 0.0659                                                                          & 0.0051          & 0.0525                                                                         & 0.0022          & 0.0120          & 0.055          & 0.193 / 0.138                                                              & 5.18 / 7.25                                                         \\
02    & 0.0659                                                                          & 0.0046          & 0.0448                                                                         & 0.0019          & 0.0110          & 0.056          & 0.182 / 0.126                                                              & 5.49 / 7.94                                                         \\
\textbf{AVG} & \textbf{0.0657}                                                                 & \textbf{0.0046} & \textbf{0.0447}                                                                & \textbf{0.0019} & \textbf{0.0107} & \textbf{0.053} & \textbf{0.181 / 0.128}                                                     & \textbf{5.52 / 7.81}                                                \\ \hline
\end{tabular}
}
\end{table*}

\section{Detailed Speed Analysis}
\OURS features 3D Gaussian Splatting~\cite{kerbl20233d} as a backbone, which allows faster rendering as it renders images directly with pointclouds instead of shooting rays from the camera along each pixel. However, our pipeline requires multiple steps that deform the gaussians as well as calculating per point deformation which slows down the rendering speed compared to the original 3D Gaussian Splatting~\cite{kerbl20233d} pipeline. Tab.~\ref{tab:time_analysis} shows detailed speed analysis of each major step. A consumer level laptop with an i9 and a RTX3080 Max-Q is used for measuring the speed.

In general, the THUman dataset~\cite{zheng2022structured} requires more time than the Mocap dataset~\cite{peng2021neural} for the animation and rendering. This is because the human performers from the Mocap dataset~\cite{peng2021neural} in general have short sleeve and short pants that require less gaussians, while the actors in the THUman dataset~\cite{zheng2022structured} wear hoodies and long pants that require more gaussians to replicate more texture, shape and shading. Furthermore, we find one of the main bottle necks is the Per Vertex Deformation (calculating the SMPL based deformation per gaussian). For the Per Vertex Deformation calculation, we use a per mesh face SVD calculation between the canonical and deformed SMPL mesh instead of using the SMPL deformation field directly for compatibility reasons, for example the Mocap dataset~\cite{peng2021neural} uses a non-conventional SMPL definition\footnote{\url{https://github.com/zju3dv/EasyMocap/blob/master/doc/02_output.md}}, while the THUman dataset~\cite{zheng2022structured}) uses the original definition. We believe that optimizing our pipeline by adapting each dataset's specific format to be compatible with the SMPL deformation field directly could speed up our pipeline up to 30\% in the future.

\begin{figure}[!t]
 \centering
    \includegraphics[width=1.0\linewidth]{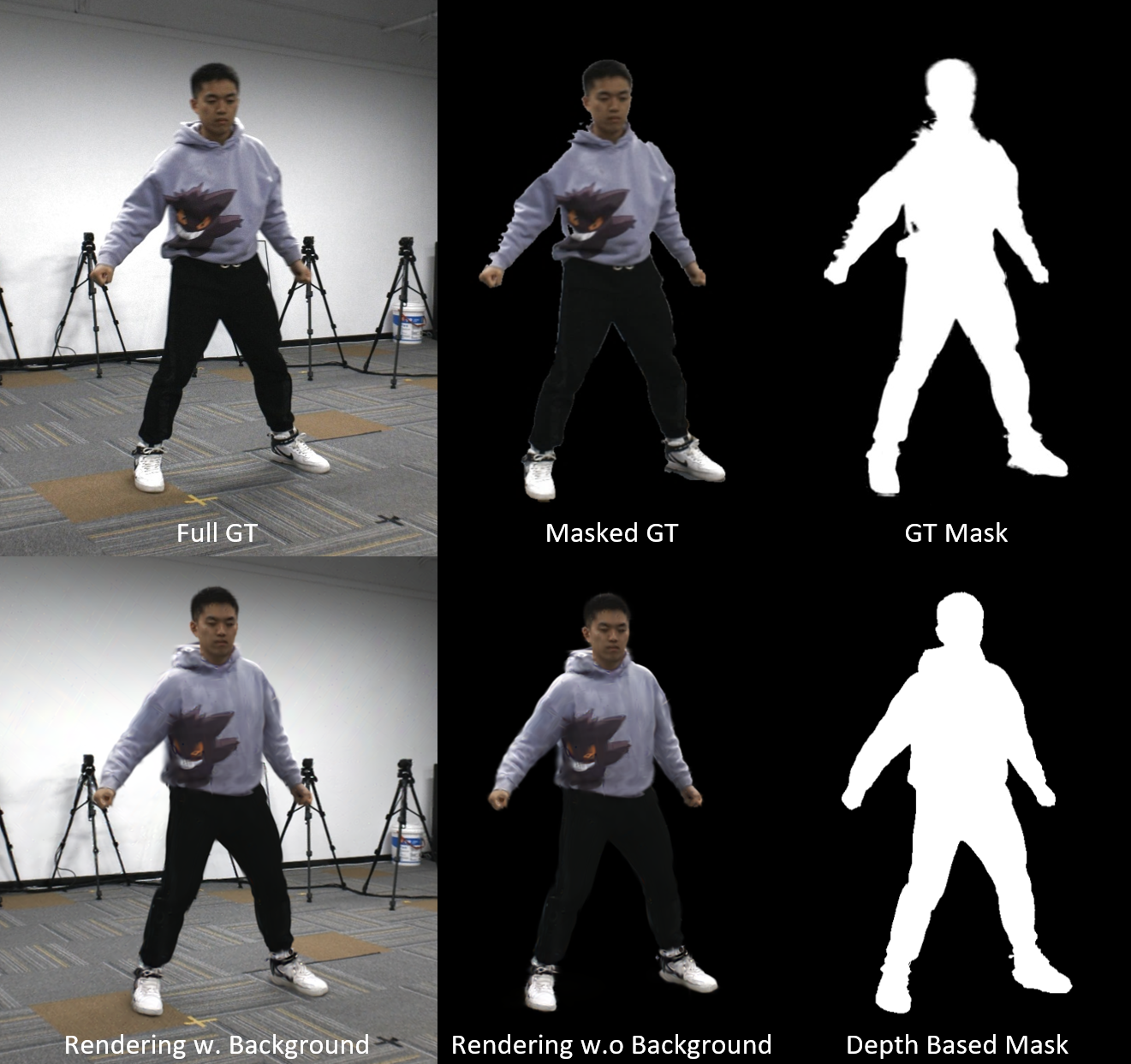}
    \caption{\textbf{Mask Comparison.} \OURS pipeline learns to build avatars together with the background to avoid any artifact related to mask annotations. During the inference the background can easily be filtered out so that the avatar can be rendered without any background.}
    \label{fig:mask_comparision}
\end{figure}

\begin{figure}[!t]
 \centering
    \includegraphics[width=1.0\linewidth]{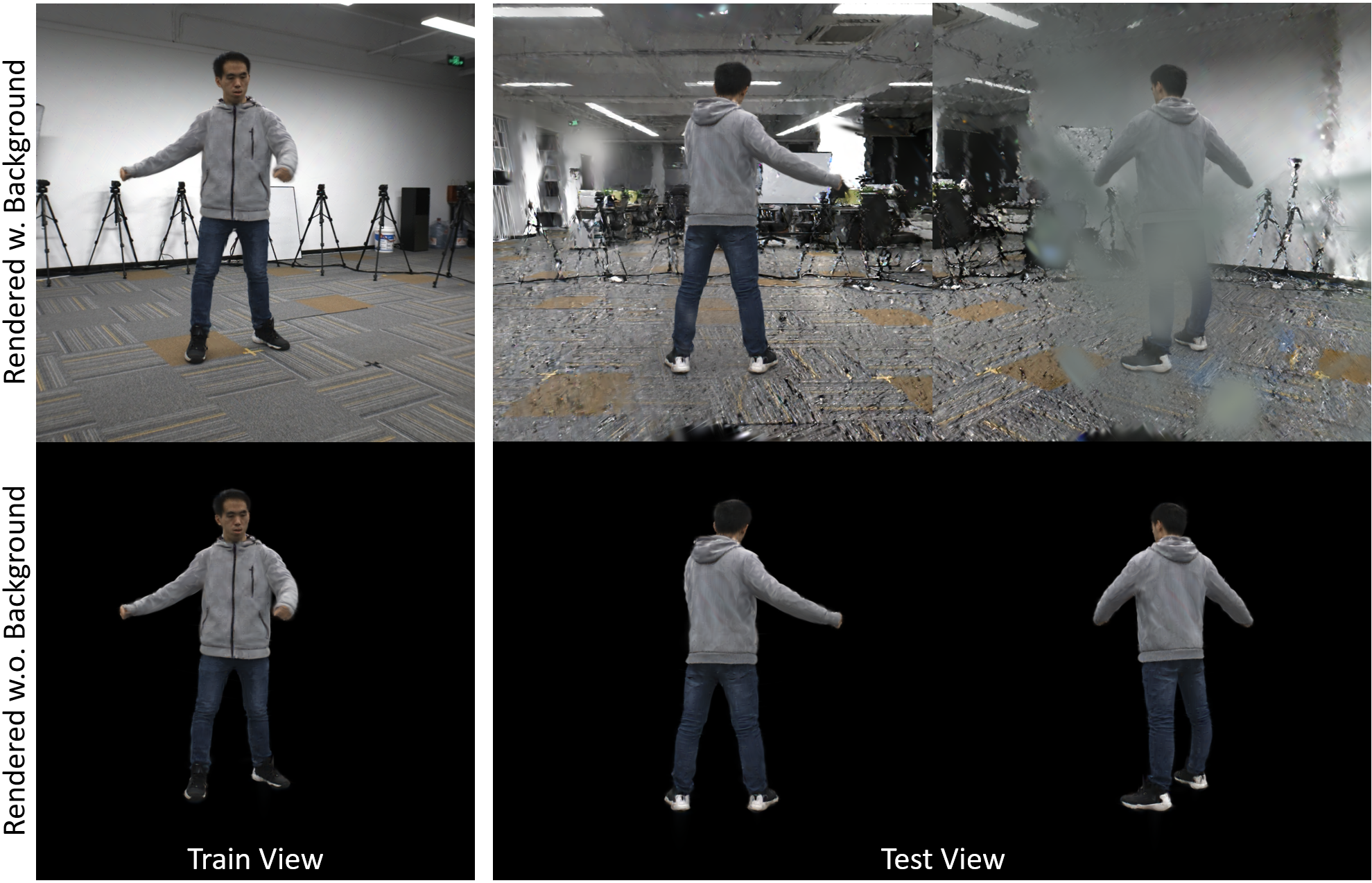}
    \caption{\textbf{Rendering without background.} As our pipeline trains with less camera views, we initialize the background gaussians with rather random points on a spherical surface around the world center instead of using SfM points~\cite{schoenberger2016sfm, schoenberger2016mvs}. With this, the learning of the background is done by overfitting on the training views, creating artifacts on unseen camera views (first row). However, this artifacts can be easily resolved by filtering the background gaussians using their parent information (second row).}
    \label{fig:rendering_with_background}
\end{figure}

\section{Rendering Examples}
In this section, we show more examples of rendered images, such as mask rendering and rendering without background in different camera poses. 

\subsection{Mask Comparison}
Annotating masks of a human performer in a long sequence is often challenging, especially when the background is uncontrolled. We show in Fig.~\ref{fig:camera_issue} show that masks on THUman 4.0~\cite{zheng2022structured} often over-segment and under-segment depending on the light condition in the frame and show in Fig.~\ref{fig:ablation_mask} that this issue can create strong artifacts when these masks are used during avatar generation. Here in Fig.~\ref{fig:mask_comparision}, we show that ours trained without masks can generate better quality masks than what is provided as GT, thus shows a big advantage of our work that can circumvent the issue of annotating good quality masks and still learns the shape of a human in a good quality. To generate the mask, we first remove the background gaussians, then render the depth by applying some depth rendering modification to the rendering script of Ziyi Yang\footnote{\url{https://github.com/graphdeco-inria/diff-gaussian-rasterization/pull/3}} followed by masking with a distance threshold.

\subsection{Rendering without Background}
For training, we initialize the background gaussians as a random pointcloud along a spherical surface instead of using a SfM~\cite{schoenberger2016sfm, schoenberger2016mvs} pointcloud, such that our pipeline stays simple and can run with fewer viewpoints (i.e. 7 views or 1 view). For this reason, the background overfits on the training views and produces artifacts in the rendering, such as the breaking of background geometry as well as generating floating points that block the camera. Fig.~\ref{fig:rendering_with_background} (first row) shows a visualization of the aforementioned artifacts. These artifacts can be easily removed by filtering the background using the parent information of the gaussians (Fig.~\ref{fig:rendering_with_background}, second row). We believe that a monocular video sequence that moves around an human performer could allow a SfM~\cite{schoenberger2016sfm, schoenberger2016mvs} pipeline to generate a pointcloud of the background and give better results for the background, allowing the rendering of novel views of the human avatar together with the background.

% \section{Rationale}
% \label{sec:rationale}
% % 
% Having the supplementary compiled together with the main paper means that:
% % 
% \begin{itemize}
% \item The supplementary can back-reference sections of the main paper, for example, we can refer to \cref{sec:intro};
% \item The main paper can forward reference sub-sections within the supplementary explicitly (e.g. referring to a particular experiment); 
% \item When submitted to arXiv, the supplementary will already included at the end of the paper.
% \end{itemize}
% % 
% To split the supplementary pages from the main paper, you can use \href{https://support.apple.com/en-ca/guide/preview/prvw11793/mac#:~:text=Delete%20a%20page%20from%20a,or%20choose%20Edit%20%3E%20Delete).}{Preview (on macOS)}, \href{https://www.adobe.com/acrobat/how-to/delete-pages-from-pdf.html#:~:text=Choose%20%E2%80%9CTools%E2%80%9D%20%3E%20%E2%80%9COrganize,or%20pages%20from%20the%20file.}{Adobe Acrobat} (on all OSs), as well as \href{https://superuser.com/questions/517986/is-it-possible-to-delete-some-pages-of-a-pdf-document}{command line tools}.

% {
%     \small
%     \bibliographystyle{ieeenat_fullname}
%     \bibliography{main}
% }

\end{document}